\documentclass[runningheads]{llncs}
\usepackage{graphicx}
\usepackage{amsmath}
\usepackage{float}
\usepackage{listings}
\usepackage{color, colortbl}
\usepackage{mathtools}
\usepackage[ruled,vlined]{algorithm2e}
\usepackage[T1]{fontenc}
\usepackage{mathtools}
\usepackage{tikz}
\usetikzlibrary{shapes,decorations,arrows,calc,arrows.meta,fit,positioning}
\tikzset{
    -Latex,auto,node distance =2 cm and 2 cm,semithick,
    state/.style ={ellipse, draw, minimum width = 1.2 cm},
    point/.style = {circle, draw, inner sep=0.09cm,fill,node contents={}},
    bidirected/.style={Latex-Latex,dashed},
    el/.style = {inner sep=2pt, align=left, sloped}
}
\usepackage{tikz}

\usepackage{makecell}

\usepackage{amsmath,amssymb} 

\usepackage{etoolbox}

\usepackage[inline]{enumitem}
\makeatletter
\newcommand{\inlineitem}[1][]{%
\ifnum\enit@type=\tw@
    {\descriptionlabel{#1}}
  \hspace{\labelsep}
\else
  \ifnum\enit@type=\z@
       \refstepcounter{\@listctr}\fi
    \quad\@itemlabel\hspace{\labelsep}
\fi}
\makeatother
\newtheorem{exmp}{Example} 

\usepackage{textalpha}
\DeclareUnicodeCharacter{3B1}{\ensuremath{\alpha}}

\usepackage{multirow}

\begin{document}
\title{Machine Learning and Constraint Programming for Efficient Healthcare Scheduling}
\titlerunning{CP and ML for Efficient Healthcare Scheduling}
%
\author{Aymen Ben Said \and Malek Mouhoub}
\authorrunning{A. Ben Said et al.}
%
\institute{Department of Computer Science, University of Regina, SK, Canada, \email{\{aymenbensaid,mouhoubm\}@uregina.ca}}

\maketitle              
\begin{abstract}
Solving combinatorial optimization problems involve satisfying a set of hard constraints while optimizing some objectives. In this context, exact or approximate methods can be used. While exact methods guarantee the optimal solution, they often come with an exponential running time as opposed to approximate methods that trade the solution’s quality for a better running time. In this context, we tackle the Nurse Scheduling Problem (NSP). The NSP consist in assigning nurses to daily shifts within a planning horizon such that workload constraints are satisfied while hospital’s costs and nurses’ preferences are optimized. To solve the NSP, we propose implicit and explicit approaches. In the implicit solving approach, we rely on Machine Learning methods using historical data to learn and generate new solutions through the constraints and objectives that may be embedded in the learned patterns. To quantify the quality of using our implicit approach in capturing the embedded constraints and objectives, we rely on the Frobenius Norm, a quality measure used to compute the average error between the generated solutions and historical data. To compensate for the uncertainty related to the implicit approach given that the constraints and objectives may not be concretely visible in the produced solutions, we propose an alternative explicit approach where we first model the NSP using the Constraint Satisfaction Problem (CSP) framework. Then we develop Stochastic Local Search methods and a new Branch and Bound algorithm enhanced with constraint propagation techniques and variables/values ordering heuristics. Since our implicit approach may not guarantee the feasibility or optimality of the generated solution, we propose a data-driven approach to passively learn the NSP as a constraint network. The learned constraint network, formulated as a CSP, will then be solved using the methods we listed earlier.

\keywords{Nurse Scheduling Problem \and Constraint Programming  \and Machine Learning \and Combinatorial Optimization}
\end{abstract}
\section{Introduction} 
Combinatorial optimization problems play a significant role in various industry applications. Solving these problems involves finding the optimal solution among feasible solutions for many real-world problems. Leveraging combinatorial optimization methods in applications like scheduling can effectively optimize resource management costs by efficient personnel scheduling and improve the overall decision-making processes. In this context, we tackle the \textit{Nurse Scheduling Problem} (NSP). Solving the NSP consists of assigning nurses to appropriate shifts satisfying a set of constraints while optimizing hospital costs and/or nurses' preferences that may be obtained from the nurses over a given planning horizon. Many methods and approaches from the areas of Constraint Programming (CP) and Operation Research (OR) were proposed to solve combinatorial optimization problems ranging from exact and approximate methods. While exact methods are able to find the optimal solution for a given problem, they often suffer from their exponential running time cost, especially for large-size problem instances with respect to the number of variables and domain size \cite{woeginger2003exact}. 

Approximate methods such as metaheuristic and Stochastic Local Search (SLS) may be considered in this regard as they are known to relatively trade the quality of the solution over the execution running time. Most of the metaheuristic methods start by randomly generating a population of candidate solutions and then try to improve the solutions by transitioning between exploration and exploitation using some type of parameters/heuristics and relying on a fitness function to measure the goodness of the solution. However, the main challenge with metaheuristic methods is tuning the parameters/heuristics to find a good balance between exploration and exploitation in order to avoid local minima/maxima \cite{RePEc:spr:snopef:v:2:y:2021:i:3:d:10.1007_s43069-021-00068-x}. SLS methods follow some type of a randomized process and often use a greedy approach for decision-making while searching for a solution. Lately, there has been an interest in utilizing Machine Learning (ML) to support solving combinatorial optimization problems that may be represented with historical data \cite{B}. The latter is achieved by learning a model through analyzing and examining the patterns among historical data and then attempting to simulate or predict similar solutions. One of the main challenges in explicitly solving combinatorial problems is the modeling phase as the constraints may not usually be available due to privacy and confidentiality concerns. However, previous solutions that were created manually exist. Therefore, ML methods may be explored as an alternative in this regard to implicitly address the constraints unavailability limitation, as well as tackle the parameters tuning issue in approximate methods and the time complexity challenge in exact methods. This submission is an extension to our conference papers \cite{NSPML,NSPCSP,10.1007/978-3-031-34020-8_5,icores24}. This paper summarizes our contributions in proposing implicit and explicit approaches to solve the NSP, provides practical examples of applying our two solving approaches in real-world scenarios, reviews previous constraints learning methodologies, and further suggests passive learning methods which can be used as an initial phase in modeling the NSP for our explicit solving approach.

\section{Literature Review}
Numerous approaches have been proposed to solve the NSP \cite{maenhout2010branching,abdennadher1999nurse,jaumard1998generalized,legrain2020rotation}. In addition to solving the NSP using exact methods, researchers have been proposing evolutionary methods based on meta-heuristics that works by eliciting candidate solutions while balancing between exploration and exploitation to escape local minima/maxima \cite{jan2000evolutionary,gutjahr2007aco,wu2013ant,jafari2015maximizing,rajeswari2017directed}. These approximate methods may be used as an alternative to exact methods to overcome the unpractical processing time, however, they do not guarantee the optimal solution. This section surveys existing work related to the modeling and solving of the NSP. More precisely, the exact and approximate methods used for the solving phase, and the active and passive methods for the modeling phase of the NSP. Kumar et al. \cite{KumarMohit2021DCSP} proposed a spreadsheet plug-in tool named SynthCSP to model and solve the NSP using a single model. SynthCSP relies on two sub-tools; COUNT-OR \cite{kumar}, and TaCLe \cite{tacle}. The latter works by explicitly learning constraints from input spreadsheet tables, and then uses the learned constraints to automatically solve partially completed schedules. The limitation of SynthCSP is that it only learns workload constraints and does not consider other objectives (like nurses' preferences) that may optimize the solution. Kumar et al. \cite{E} also proposed an explicit learning method named ARNOLD to partially automates the modelling process using a constraint language. ARNOLD is slightly different than COUNT-OR \cite{kumar} because it requires the quantities of interest (such as tensor' lower/upper bounds) to be given manually by the user instead of learning them automatically from the schedules. Branch \& Bound (B\&B) was explored in \cite{baskaran2014integer} via integer programming to solve the NSP. However, exact methods such as B\&B suffer from their exponential running time cost, especially when it comes to solving large-size problem instances with larges number of variables and domain size \cite{woeginger2003exact}. Hybrid methods that combine more than one algorithm were considered to solve the NSP \cite{burke2001memetic}. For instance, the authors in \cite{zhang2011hybrid} proposed a hybrid method that combined both Genetic Algorithm (GA) and Variable Neighborhood Search (VNS). GA was used to solve sub-problems and return initial feasible solutions that are fed into VNS to improve them. Constantino et al. \cite{I} proposed a hybrid heuristic algorithm aiming to maximize the preference of each nurse individually. The latter is a two-phased algorithm, the first phase produces initial shift patterns, and the second phase involves conducting some shift reassignments to balance the nurses' preferences. Constantino et al. \cite{J} proposed another variant of this algorithm relying on VNS in the second phase to improve the initially obtained solution instead of using local search. Tassopoulos et al. \cite{Tassopoulos} proposed a meta-heuristic method using a two-phase VNS algorithm to solve the NSP. Experimental results using multiple INRC-2010 competition instances revealed promising results in terms of maximizing the objective function. ilmaz et al. \cite{YilmazEbru2010AMPM} proposed a mathematical model that focuses on minimizing nurses' idle wait time during the planning horizon. The authors evaluated the model using a solver named LINGO8.0 and the experimentation revealed promising results in terms of ensuring a globally optimal solution. Aickelin et al. \cite{li,Aickelin_2007} proposed a new Bayesian networks based scheduling algorithm. The algorithm aim to mimic the human scheduler' explicit learning by constructing a Bayesian network that represents the joint distribution of solutions. The experimental results from real-world NSP data demonstrated the effectiveness of the proposed algorithm. Snehasish et al. \cite{H} solved the NSP explicitly using multiple meta-heuristic methods; Firefly Algorithm, Particle Swarm Optimization (PSO), Simulated Annealing, and Genetic Algorithm. The experimentation consisted of comparing the four methods in terms of quality of the solution and it showed that the Firefly Algorithm performed the best among all methods due to its randomness nature that triggers more exploitation in the search space, while PSO had the worst performance overall. Most of the above-mentioned methods in this section \cite{li,H,I,J} requires the constraints and objectives to be explicitly defined and the solving phase is conducted using exact and approximate methods. Approximate methods has the time advantage but come with a degree of uncertainty related to the quality of the solution while exact methods guarantee the optimal solution but may come with a time complexity trade-off. Various methodologies have been proposed in the area of Constraints Learning to simplify the task of modeling combinatorial problems. This section surveys existing works related to active and passive modeling methods for combinatorial problems including the NSP. The generate-and-test approach is among the popular approaches used for learning constraints and it consist in generating all possible constraints and verifying their satisfiability by querying feasible and/or infeasible solutions. Bessiere et al. \cite{BESSIERE2017315} proposed passive and active learning algorithms namely ``CONACQ1" and ``CONACQ2", respectively. CONACQ1 automates the learning of constraint network using both positive and negative examples, while CONACQ2 relies on membership queries to classify examples. A major concern of learning constraints using membership queries is that the number of queries needed to converge may be large \cite{DBLP:conf/ijcai/AlanaziMZ16,DBLP:journals/ai/AlanaziMZ20,mouhoub2023exact}. An alternative in this regard consist of using partial queries \cite{BESSIERE2023103896} considering a subset of variables rather than all the variables. Sergey et al. \cite{tacle} proposed a constraint learner system based on a tabular structure (precisely spreadsheet) which may be considered to learn constraints in the NSP context since schedules are represented in a similar structure. ModelSeeker \cite{modelseeker} is another method for learning global constraints from positive examples through a predefined constraint catalog. Kumar et al.  \cite{kumar} proposed a method called ``COUNT-OR" to automate the learning of NSP constraints. ``COUNT-OR" uses historical data and relies on CP and ML based techniques. ``COUNT-OR" was evaluated in terms of efficiency of learning constraints using multiple solutions from the NRC-II competition \cite{ceschia2015second}. The constraints acquisition of the latter is achieved using matrix operations for the purpose of learning different quantities of interest (e.g. bounds related to working nurses per day) then using these quantities to build new solutions. One limitation of COUNT-OR is that it is restricted to a specific instance dimension rather than a general one. In the same context, ``COUNT-CP" \cite{lirias3973866} was proposed as an extension to ``COUNT-OR" to learn first-order constraints regardless of the instances dimension. 

\section{Implicit Solving of the NSP}
\label{implicit}
We propose an implicit solving approach based on ML methods (Association Rules Mining , High Utility Item-set Mining, Naive Bayes, and Bayesian Network) and historical data without any prior knowledge. The aim of our solving approach is to implicitly learn the frequent patterns from historical data in the form of association rules and ML models, and then use the learned patterns that embeds the constraints and objectives to simulate new scheduling solutions. Note that historical data may encompass many types of constraints and objectives including workload constraints, financial constraints, conflicts between staff, etc. Thus, our implicit approach not only solves the NSP but also maintains the integrity and confidentiality of the data. The proposed methods were evaluated, and the results were reported in terms of closeness to the input data using the Frobenius Norm. 

\subsection{Frequent Pattern Mining via Apriori}
Association rules mining \cite{F} is an unsupervised learning method for discovering frequent patterns and associations between items in transactional databases. Association rules mining was mainly introduced in the market basket analysis to analyze and learn the itemsets that are frequently purchased together in the form of association rules such as ``item1 \& item2 $\rightarrow$ item3'' which translates to: if item1 and item2 occur together, item3 is highly likely to also occur. The items in the NSP context are the nurses, therefore, we adopt the Apriori algorithm \cite{F,10.5555/645920.672836} to learn associations between nurses. More precisely, we apply Apriori on historical scheduling solutions to extract the frequent assignments of nurses using interestingness measures; support and confidence, under user-defined thresholds. The Apriori algorithm relies on a powerful property to prune the infrequent itemsets and minimize the search space. The Apriori property states that all the subsets of a frequent item-set should be frequent, and if an item-set is not frequent, then all its super-sets will be infrequent \cite{10.5555/645920.672836}. The generated rules implicitly represent the assignments that could be enforced by the constraints and/or preferences, and are utilized to simulate new scheduling scenarios. 

\begin{exmp}
Let us consider the transactions table in Table. \ref{transactionstable} (which may be obtained by preprocessing a regular schedule). The target is to find frequent nurse assignments and generate association rules with the following parameters thresholds: min-support: 2 and min-confidence: 60\%. 

\begin{table}[ht]
\centering
\caption{Transactions Table} 
\label{transactionstable}
	\begin{tabular}{ | c | c |} \hline
        Day/Transaction & Items\\[1ex] \hline
        $Day_{1}$ & $Nurse_2, Nurse_4$ \\[1ex] \hline
		$Day_{2}$ & $Nurse_1, Nurse_3, Nurse_5$ \\[1ex] \hline
		$Day_{3}$ & $Nurse_1, Nurse_2, Nurse_4, Nurse_5$ \\[1ex] \hline
		$Day_{4}$ & $Nurse_2, Nurse_3, Nurse_4$ \\[1ex] \hline
		$Day_{5}$ & $Nurse_1, Nurse_2, Nurse_5$ \\[1ex] \hline
		$Day_{6}$ & $Nurse_3, Nurse_4$ \\[1ex] \hline
        $Day_{7}$ & $Nurse_1, Nurse_3, Nurse_4, Nurse_5$ \\[1ex] \hline
	\end{tabular}
\end{table}

\noindent The process of generating the association rules from the transactional database in Table \ref{transactionstable} consist of computing the support counts of all the items (Nurses) from the transactions table, finding all frequent super-sets with two, three, four, etc. items and prune the infrequent ones based on the minimum support threshold. After discovering all the frequent item-sets, the final step consist of generating all the association rules and computing their confidence values. Note that the association rules found from the transactional database in Table \ref{transactionstable} based on min-support count of 2 are: \{$Nurse_1, Nurse_2, Nurse_5$\}, \{$Nurse_1, Nurse_3, Nurse_5$\}, \{$Nurse_1, Nurse_4, Nurse_5$\}, and the rules with min-confidence of 60\% that will be used to simulate new schedules are listed below.
\begin{enumerate}
\item $Nurse_1 \ \& \ Nurse_2 \Rightarrow Nurse_5$
\item $Nurse_2 \ \& \ Nurse_5 \Rightarrow Nurse_1$
\item $Nurse_1 \ \& \ Nurse_3 \Rightarrow Nurse_5$
\item $Nurse_1 \ \& \ Nurse_5 \Rightarrow Nurse_3$
\item $Nurse_3 \ \& \ Nurse_5 \Rightarrow Nurse_1$
\item $Nurse_1 \ \& \ Nurse_4 \Rightarrow Nurse_5$
\item $Nurse_4 \ \& \ Nurse_5 \Rightarrow Nurse_1$
\end{enumerate}
\end{exmp}

\subsection{High Utility Itemset Mining via Two-Phase}
High Utility Item-set Mining (HUIM) \cite{F} is derived from the mining framework. Given that Apriori is limited to only considering the frequency of itemsets rather than the utility, we rely on the Two-Phase algorithm \cite{twophase} to overcome this limitation and learn the itemsets that maximize the utility in addition to the frequency. By doing so, we aim to extract more relevant itemsets and construct more accurate scheduling solutions. The input of the Two-Phase algorithm consist of workload coverage and the nurses' average preferences in the form of external utility table. Note that the utility table in the context of the NSP are the preference costs related to nurses working in specific shifts which we obtain from the NSPLib instances \cite{lib}. Similar to Apriori, Two-Phase uses a min-utility threshold to elicit the high utility itemsets. Setting the min-utility parameter is very challenging as stated in literature \cite{M} and may require domain knowledge. For simplicity, we set a high value for this parameter with respect to the number of generated high utility itemsets since we are interested in item-sets of high utilities. 

\begin{exmp}
Let us consider the transaction table given in Table. \ref{HUIMinputschedule} and the utility table given in Table. \ref{utilitytable} representing the record of workload coverage of shift 1 in a given month, and the nurses' preferences over shift 1, respectively. The goal is to find the high-utility itemsets that maximize the overall nurses' preferences in working shift 1 during the schedule. 

\begin{table}[H]
\begin{minipage}{0.6\textwidth}
  \centering
  \caption{Transaction table (T)} 
  \label{HUIMinputschedule}
  \begin{tabular}{ | c | c | c| c| c| c | c | c | c | c |}
    \hline
    & $Nurse_1$ & $Nurse_2$ & $Nurse_3$ & $Nurse_4$ & $Nurse_5$ \\[1ex] \hline
    $Day_1Shift_1$ & 0 & 1 & 0 & 3 & 0 \\[1ex] \hline
    $Day_2Shift_1$ & 2 & 0 & 2 & 0 & 2 \\[1ex] \hline
    $Day_3Shift_1$ & 1 & 3 & 0 & 2 & 1 \\[1ex] \hline
    $Day_4Shift_1$ & 0 & 1 & 4 & 1 & 0 \\[1ex] \hline
    $Day_5Shift_1$ & 3 & 2 & 0 & 0 & 2 \\[1ex] \hline
    $Day_6Shift_1$ & 0 & 0 & 1 & 0 & 0 \\[1ex] \hline
    $Day_7Shift_1$ & 1 & 1 & 0 & 1 & 1 \\[1ex] \hline
  \end{tabular}
\end{minipage}%
\begin{minipage}{0.5\textwidth}
  \centering
  \caption{Utility table (U)} 
  \label{utilitytable}
  \begin{tabular}{ | c | c |}
    \hline
    $Items$ & $Preferences$  \\[1ex] \hline
    $Nurse_1$ & 3 \\[1ex] \hline
    $Nurse_2$ & 4 \\[1ex] \hline
    $Nurse_3$ & 2 \\[1ex] \hline
    $Nurse_4$ & 3 \\[1ex] \hline
    $Nurse_5$ & 1 \\[1ex] \hline
  \end{tabular}
\end{minipage}
\end{table}

The transaction utility table is depicted in Table. \ref{transactionutility} and it indicates the preference score achieved on each day of the transaction table. It is obtained by computing the dot product of the transaction table and the utility table as shown below.

\[
T \cdot U = 
\begin{bmatrix}
0 & 1 & 0 & 3 & 0 \\
2 & 0 & 2 & 0 & 2 \\
1 & 3 & 0 & 2 & 1 \\
0 & 1 & 4 & 1 & 0 \\
3 & 2 & 0 & 0 & 2 \\
0 & 0 & 1 & 0 & 0 \\
1 & 1 & 0 & 1 & 1 \\
\end{bmatrix} \hfill
\cdot
\begin{bmatrix}
3 \\
4 \\
2 \\
3 \\
1 \\
\end{bmatrix}
=
\begin{bmatrix}
13 \\
12 \\
22 \\
15 \\
19 \\
2 \\
11 \\
\end{bmatrix}
\]

\begin{table}[H]
\centering
\caption{Transaction utility} 
\label{transactionutility}
	\begin{tabular}{ | c | c |}
 		\hline
 	TID & TU  \\[1ex] \hline
 	$Day_1Shift_1$ & 13 \\[1ex] \hline
		$Day_2Shift_1$ & 12 \\[1ex] \hline
		$Day_3Shift_1$ & 22 \\[1ex] \hline
		$Day_4Shift_1$ & 15 \\[1ex] \hline
		$Day_5Shift_1$ & 19 \\[1ex] \hline
        $Day_6Shift_1$ & 2 \\[1ex] \hline
		$Day_7Shift_1$ & 11 \\[1ex] \hline
	\end{tabular}
\end{table}

The Two-Phase algorithm relies on the Transaction Weighted Utilization (TWU) upper bound to over-estimate the high utility itemsets and uses the Transaction-weighted Downward Closure Property to prune the search space in Phase I; given an itemset X, the TWU of a superset of itemset X cannot be greater than the TWU of itemset X, so any low transaction weighted utilization itemset cannot have a high transaction weighted utilization superset. The TWU is simply the sum of the transaction utilities of all the transactions that contain a given itemset. For example, the itemset \{$Nurse_{1}Nurse_{2}$\} appears in transactions; $Day_3Shift_1$, $Day_5Shift_1$, and $Day_7Shift_1$, thus,  TWU($Nurse_{1}Nurse_{2}$)= 22+19+11=52. An itemset is considered a high transaction weighted utilization only if its TWU is greater or equal to a minimum threshold $\varepsilon$’ set by the user. Two-Phase assumes $\varepsilon$’ to be equal to $\varepsilon$ (minimum utility) to guarantee all the high utility itemsets to be included in the set of high transaction weighted utilization itemsets found in Phase I. Phase II consists of scanning the transaction table to filter out the low utility itemsets from the high transaction weighted utilization itemsets obtained in Phase I by computing their real utility. Figure. \ref{huimsearchspace} depicts the Two-Phase search space where the tuples under the itemsets in the rectangles are the TWU/Nº of Occurrences of the itemsets. The following notation is used to elicit the itemsets; $N_{1}: Nurse_{1}, N_{2}: Nurse_{2}, N_{3}: Nurse_{3}, N_{4}: Nurse_{4}, N_{5}: Nurse_{5}$. Assuming $\varepsilon$’ = $\varepsilon$ = 15, Phase I yields 19 high transaction-weighted utilization itemsets: \{$N_{1}$\}, \{$N_{2}$\}, \{$N_{3}$\}, \{$N_{4}$\}, \{$N_{5}$\}, \{$N_{1}N_{2}$\}, \{$N_{1}N_{4}$\}, \{$N_{1}N_{5}$\}, \{$N_{2}N_{3}$\}, \{$N_{2}N_{4}$\}, \{$N_{2}N_{5}$\}, \{$N_{3}N_{4}$\}, \{$N_{4}N_{5}$\}, \{$N_{1}N_{2}N_{4}$\}, \{$N_{1}N_{2}N_{5}$\}, \{$N_{1}N_{4}N_{5}$\}, \{$N_{2}N_{3}N_{4}$\}, \{$N_{2}N_{4}N_{5}$\}, \{$N_{1}N_{2}N_{4}N_{5}$\}. After computing the real utilities and filtering out the low utility itemsets in Phase II (as shown below), we obtain the following 14 high utility itemsets: \{$N_{1}$\}, \{$N_{2}$\}, \{$N_{4}$\}, \{$N_{1}N_{2}$\}, \{$N_{1}N_{4}$\}, \{$N_{1}N_{5}$\}, \{$N_{2}N_{4}$\}, \{$N_{2}N_{5}$\}, \{$N_{1}N_{2}N_{4}$\}, \{$N_{1}N_{2}N_{5}$\}, \{$N_{1}N_{4}N_{5}$\}, \{$N_{2}N_{3}N_{4}$\}, \{$N_{2}N_{4}N_{5}$\}, \{$N_{1}N_{2}N_{4}N_{5}$\}. \\
Computing the utilities of itemsets:\\
\{$N_{1}\} = (2 \cdot 3) + (1 \cdot 3) + (3 \cdot 3) + (1 \cdot 3) = 21$ \\
\{$N_{2}\} = (1 \cdot 4) + (3 \cdot 4) + (1 \cdot 4) + (2 \cdot 4) + (1 \cdot 4) = 32$ \\
\{$N_{3}\} = (2 \cdot 2) + (4 \cdot 2) + (1 \cdot 2) = 14$ \\
\{$N_{4}\} = (3 \cdot 3) + (2 \cdot 3) + (1 \cdot 3) + (1 \cdot 3) = 21$ \\
\{$N_{5}\} = (2 \cdot 1) + (1 \cdot 1) + (2 \cdot 1) + (1 \cdot 1) = 6$ \\
\{$N_{1}N_{2}\} = (1 \cdot 3 + 3 \cdot 4) + (3 \cdot 3 + 2 \cdot 4) + (1 \cdot 3 + 1 \cdot 4) = 39$ \\
\{$N_{1}N_{4}\} = (1 \cdot 3 + 2 \cdot 3) + (1 \cdot 3 + 1 \cdot 3) = 15$ \\ 
\{$N_{1}N_{5}\} = (2 \cdot 3 + 2 \cdot 1) + (1 \cdot 3 + 1 \cdot 1) + (3 \cdot 3 + 2 \cdot 1) + (1 \cdot 3 + 1 \cdot 1) = 27$ \\
\{$N_{2}N_{3}\} = (1 \cdot 4 + 4 \cdot 2) = 12$ \\ 
\{$N_{2}N_{4}\} = (1 \cdot 4 + 3 \cdot 3) + (3 \cdot 4 + 2 \cdot 3) + (1 \cdot 4 + 1 \cdot 3) + (1 \cdot 4 + 1 \cdot 3) = 45$ \\
\{$N_{2}N_{5}\} = (3 \cdot 4 + 1 \cdot 1) + (2 \cdot 4 + 2 \cdot 1) + (1 \cdot 4 + 1 \cdot 1) = 28$ \\ 
\{$N_{3}N_{4}\} = (4 \cdot 2 + 1 \cdot 3) = 11$ \\ 
\{$N_{4}N_{5}\} = (2 \cdot 3 + 1 \cdot 1) + (1 \cdot 3 + 1 \cdot 1) = 11$ \\ 
\{$N_{1}N_{2}N_{4}\} = (1 \cdot 3 + 3 \cdot 4 + 2 \cdot 3) + (1 \cdot 3 + 1 \cdot 4 + 1 \cdot 3) = 31$ \\  
\{$N_{1}N_{2}N_{5}\} = (1 \cdot 3 + 3 \cdot 4 + 1 \cdot 1) + (3 \cdot 3 + 2 \cdot 4 + 2 \cdot 1) + (1 \cdot 3 + 1 \cdot 4 + 1 \cdot 1) = 43$ \\ 
\{$N_{1}N_{4}N_{5}\} = (1 \cdot 3 + 2 \cdot 3 + 1 \cdot 1) + (1 \cdot 3 + 1 \cdot 3 + 1 \cdot 1) = 17$ \\
\{$N_{2}N_{3}N_{4}\} = (1 \cdot 4 + 4 \cdot 2 + 1 \cdot 3) = 15$ \\
\{$N_{2}N_{4}N_{5}\} = (3 \cdot 4 + 2 \cdot 3 + 1 \cdot 1) + (1 \cdot 4 + 1 \cdot 3 + 1 \cdot 1) = 27$ \\
\{$N_{1}N_{2}N_{4}N_{5}\} = (1 \cdot 3 + 3 \cdot 4 + 2 \cdot 3 + 1 \cdot 1) + (1 \cdot 3 + 1 \cdot 4 + 1 \cdot 3 + 1 \cdot 1) = 33$   

\begin{figure}[ht]
\begin{center}
\resizebox{\linewidth}{!}{%
\begin{tikzpicture}[node distance=2.4cm, every node/.style={rectangle, draw, font=\footnotesize, align=center}]
\node (l1-1) at (0,0) {$N_{1}N_{2}N_{3}N_{4}N_{5}$ \\ 0/0};
\node (l2-1) at (-6,-2.5) {$N_{1}N_{2}N_{3}N_{4}$ \\ 0/0};
\node (l2-2) at (-3,-2.5) {$N_{1}N_{2}N_{3}N_{5}$ \\ 0/0};
\node (l2-3) at (0,-2.5) {$N_{1}N_{2}N_{4}N_{5}$ \\ 33/2};
\node (l2-4) at (3,-2.5) {$N_{1}N_{3}N_{4}N_{5}$ \\ 0/0};
\node (l2-5) at (6,-2.5) {$N_{2}N_{3}N_{4}N_{5}$ \\ 0/0};
\node (l3-1) at (-9,-5) {$N_{1}N_{2}N_{3}$ \\ 0/0};
\node (l3-2) at (-7,-5) {$N_{1}N_{2}N_{4}$ \\ 33/2};
\node (l3-3) at (-5,-5) {$N_{1}N_{2}N_{5}$ \\ 52/3};
\node (l3-4) at (-3,-5) {$N_{1}N_{3}N_{4}$ \\ 0/0};
\node (l3-5) at (-1,-5) {$N_{1}N_{3}N_{5}$ \\ 12/1};
\node (l3-6) at (1,-5) {$N_{1}N_{4}N_{5}$ \\ 33/2};
\node (l3-7) at (3,-5) {$N_{2}N_{3}N_{4}$ \\ 15/1};
\node (l3-8) at (5,-5) {$N_{2}N_{3}N_{5}$ \\ 0/0};
\node (l3-9) at (7,-5) {$N_{2}N_{4}N_{5}$ \\ 33/2};
\node (l3-10) at (9,-5) {$N_{3}N_{4}N_{5}$ \\ 0/0};
\node (l4-1) at (-9,-7.5) {$N_{1}N_{2}$ \\ 52/3};
\node (l4-2) at (-7,-7.5) {$N_{1}N_{3}$ \\ 12/1};
\node (l4-3) at (-5,-7.5) {$N_{1}N_{4}$ \\ 33/2};
\node (l4-4) at (-3,-7.5) {$N_{1}N_{5}$ \\ 64/4};
\node (l4-5) at (-1,-7.5) {$N_{2}N_{3}$ \\ 15/1};
\node (l4-6) at (1,-7.5) {$N_{2}N_{4}$ \\ 64/4};
\node (l4-7) at (3,-7.5) {$N_{2}N_{5}$ \\ 52/3};
\node (l4-8) at (5,-7.5) {$N_{3}N_{4}$ \\ 15/1};
\node (l4-9) at (7,-7.5) {$N_{3}N_{5}$ \\ 12/1};
\node (l4-10) at (9,-7.5) {$N_{4}N_{5}$ \\ 33/2};
\node (l5-1) at (-6,-10) {$N_{1}$ \\ 64/4};
\node (l5-2) at (-3,-10) {$N_{2}$ \\ 83/5};
\node (l5-3) at (0,-10) {$N_{3}$ \\ 29/3};
\node (l5-4) at (3,-10) {$N_{4}$ \\ 64/4};
\node (l5-5) at (6,-10) {$N_{5}$ \\ 64/4};
\draw[-]  (l2-1.north) -- (l1-1.south);
\draw[-]  (l2-2.north) -- (l1-1.south);
\draw[-]  (l2-3.north) -- (l1-1.south);
\draw[-]  (l2-4.north) -- (l1-1.south);
\draw[-]  (l2-5.north) -- (l1-1.south);
\draw[-]  (l3-1.north) -- (l2-1.south);
\draw[-]  (l3-1.north) -- (l2-2.south);
\draw[-]  (l3-2.north) -- (l2-1.south);
\draw[-]  (l3-2.north) -- (l2-3.south);
\draw[-]  (l3-3.north) -- (l2-2.south);
\draw[-]  (l3-3.north) -- (l2-3.south);
\draw[-]  (l3-4.north) -- (l2-1.south);
\draw[-]  (l3-4.north) -- (l2-4.south);
\draw[-]  (l3-5.north) -- (l2-2.south);
\draw[-]  (l3-5.north) -- (l2-4.south);
\draw[-]  (l3-6.north) -- (l2-3.south);
\draw[-]  (l3-6.north) -- (l2-4.south);
\draw[-]  (l3-7.north) -- (l2-1.south);
\draw[-]  (l3-7.north) -- (l2-5.south);
\draw[-]  (l3-8.north) -- (l2-2.south);
\draw[-]  (l3-8.north) -- (l2-5.south);
\draw[-]  (l3-9.north) -- (l2-3.south);
\draw[-]  (l3-9.north) -- (l2-5.south);
\draw[-]  (l3-10.north) -- (l2-4.south);
\draw[-]  (l3-10.north) -- (l2-5.south);
\draw[-]  (l4-1.north) -- (l3-1.south);
\draw[-]  (l4-1.north) -- (l3-2.south);
\draw[-]  (l4-1.north) -- (l3-3.south);
\draw[-]  (l4-2.north) -- (l3-1.south);
\draw[-]  (l4-2.north) -- (l3-4.south);
\draw[-]  (l4-2.north) -- (l3-5.south);
\draw[-]  (l4-3.north) -- (l3-2.south);
\draw[-]  (l4-3.north) -- (l3-4.south);
\draw[-]  (l4-3.north) -- (l3-6.south);
\draw[-]  (l4-4.north) -- (l3-3.south);
\draw[-]  (l4-4.north) -- (l3-5.south);
\draw[-]  (l4-4.north) -- (l3-6.south);
\draw[-]  (l4-5.north) -- (l3-1.south);
\draw[-]  (l4-5.north) -- (l3-7.south);
\draw[-]  (l4-5.north) -- (l3-8.south);
\draw[-]  (l4-6.north) -- (l3-2.south);
\draw[-]  (l4-6.north) -- (l3-7.south);
\draw[-]  (l4-6.north) -- (l3-9.south);
\draw[-]  (l4-7.north) -- (l3-3.south);
\draw[-]  (l4-7.north) -- (l3-8.south);
\draw[-]  (l4-7.north) -- (l3-9.south);
\draw[-]  (l4-8.north) -- (l3-4.south);
\draw[-]  (l4-8.north) -- (l3-7.south);
\draw[-]  (l4-8.north) -- (l3-10.south);
\draw[-]  (l4-9.north) -- (l3-5.south);
\draw[-]  (l4-9.north) -- (l3-8.south);
\draw[-]  (l4-9.north) -- (l3-10.south);
\draw[-]  (l4-10.north) -- (l3-6.south);
\draw[-]  (l4-10.north) -- (l3-9.south);
\draw[-]  (l4-10.north) -- (l3-10.south);
\draw[-]  (l5-1.north) -- (l4-1.south);
\draw[-]  (l5-1.north) -- (l4-2.south);
\draw[-]  (l5-1.north) -- (l4-3.south);
\draw[-]  (l5-1.north) -- (l4-4.south);

\draw[-]  (l5-2.north) -- (l4-1.south);
\draw[-]  (l5-2.north) -- (l4-5.south);
\draw[-]  (l5-2.north) -- (l4-6.south);
\draw[-]  (l5-2.north) -- (l4-7.south);

\draw[-]  (l5-3.north) -- (l4-2.south);
\draw[-]  (l5-3.north) -- (l4-5.south);
\draw[-]  (l5-3.north) -- (l4-8.south);
\draw[-]  (l5-3.north) -- (l4-9.south);

\draw[-]  (l5-4.north) -- (l4-3.south);
\draw[-]  (l5-4.north) -- (l4-6.south);
\draw[-]  (l5-4.north) -- (l4-8.south);
\draw[-]  (l5-4.north) -- (l4-10.south);

\draw[-]  (l5-5.north) -- (l4-4.south);
\draw[-]  (l5-5.north) -- (l4-7.south);
\draw[-]  (l5-5.north) -- (l4-9.south);
\draw[-]  (l5-5.north) -- (l4-10.south);

\end{tikzpicture}}
\caption{The Two-Phase algorithm search space}
\label{huimsearchspace}
\end{center}
\end{figure}
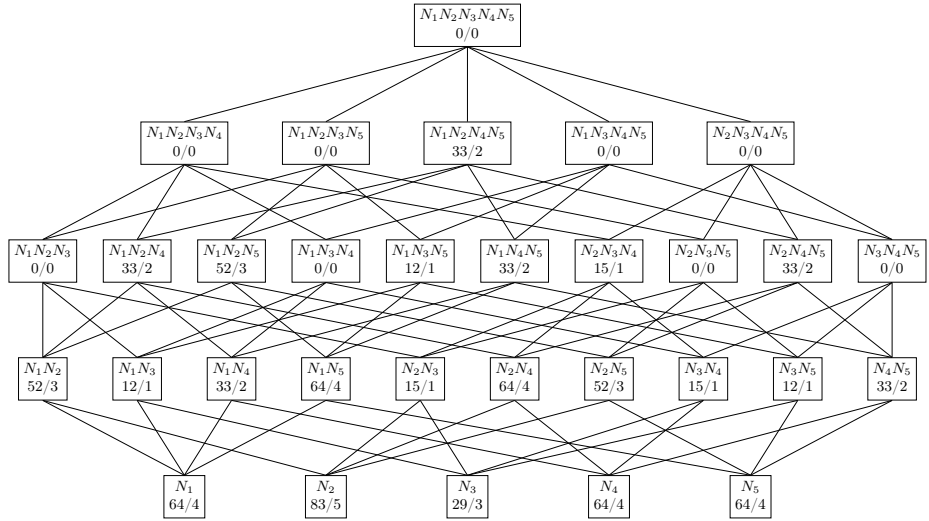

\end{exmp}

\subsection{Naive Bayes Classifier}
Naive Bayes (NB) classifiers are supervised machine learning methods derived from the Bayes Law and used for classification tasks. NB classifies new data observations of data using labeled data-set and relies on the independence assumption (i.e. naive assumption) of variables given the target class which implies that each attribute contribute independently to the target class. In the context of solving the NSP, the input schedules are expected to be partially completed and the target is to fill out the missing assignments, more precisely, we use NB to predict unknown shift assignments based on prior shift observations. 

The following is a formulation of the NSP problem according to the Bayes Law. Let $i = \{1,..,n\}$ be the nurse index, $j = \{1,..,m\}$ is the shift index.
\begin{equation*}
 N_{ij} =
    \begin{cases}
      1, & \text{if Nurse i is assigned shift j}\\
      0, & \text{otherwise}
    \end{cases}       
\end{equation*}
\begin{equation*}
P(N_{i'j'}|N_{ij}) = \frac{P(N_{ij} \cap N_{i'j'})} {P(N_{ij})}
\end{equation*}
\begin{equation*}
P(N_{ij}|N_{i'j'}) = \frac{P(N_{ij} \cap N_{i'j'})} {P(N_{i'j'})}  
\end{equation*}
\begin{equation*}
P(N_{ij} \cap N_{i'j'}) =  P(N_{ij}|N_{i'j'}) \ P(N_{i'j'})
\end{equation*}
\begin{equation}
\Rightarrow P(N_{i'j'} | N_{ij}) = \frac{P(N_{ij}|N_{i'j'}) \ P(N_{i'j'})} {P(N_{ij})} 
\end{equation}

\begin{exmp}
\label{ex2}
Let us consider the training and testing data given in Table. \ref{NBhData} and Table. \ref{NBtestData}, respectively. The target is to train a NB model and predict the nurse assignments of shift 4 for the testing data. To overcome the zero probability problem inherited by the Naive Bayes classifier, we rely on Laplace add-1 smoothing in our computation \cite{laplace}. The process of shift assignments predictions consist of computing the prior probabilities for all the class labels, computing the likelihood probability for each class, using the Bayes Law (with the computed prior and likelihood values) to calculate the posterior probabilities and enforcing Laplace add-1 smoothing to overcome the zero probability problem, and finally selecting the predicted class label with the highest probability. Below is an example of predicting the first instance of the testing data.

\begin{table}[ht]
\begin{minipage}{0.5\textwidth}
\centering
\caption{Training Data} 
	\begin{tabular}{ | c | c | c| c| c| c | c |}
 		\hline
 		& $Shift_1$ & $Shift_2$ & $Shift_3$ & $Shift_4$ \\[1ex] \hline
 	    $Day_1$ & $Nurse_1$ & $Nurse_4$ & $Nurse_3$ & $Nurse_2$  \\[1ex] \hline
		$Day_2$ & $Nurse_4$ & $Nurse_3$ & $Nurse_1$ & $Nurse_2$  \\[1ex] \hline
		$Day_3$ & $Nurse_2$ & $Nurse_1$ & $Nurse_3$ & $Nurse_4$ \\[1ex] \hline
		$Day_4$ & $Nurse_4$ & $Nurse_2$ & $Nurse_1$ & $Nurse_3$  \\[1ex] \hline
		$Day_5$ & $Nurse_3$ & $Nurse_2$ & $Nurse_4$ & $Nurse_1$  \\[1ex] \hline
		$Day_6$ & $Nurse_2$ & $Nurse_1$ & $Nurse_3$ & $Nurse_4$  \\[1ex] \hline
		$Day_7$ & $Nurse_1$ & $Nurse_2$ & $Nurse_4$ & $Nurse_3$  \\[1ex] \hline
		$Day_8$ & $Nurse_4$ & $Nurse_2$ & $Nurse_1$ & $Nurse_3$  \\[1ex] \hline
		$Day_9$ & $Nurse_2$ & $Nurse_1$ & $Nurse_4$ & $Nurse_3$ \\[1ex] \hline
		$Day_{10}$ & $Nurse_4$ & $Nurse_3$ & $Nurse_2$ & $Nurse_1$  \\[1ex] \hline
		$Day_{11}$ & $Nurse_3$ & $Nurse_1$ & $Nurse_2$ & $Nurse_4$  \\[1ex] \hline
		$Day_{12}$ & $Nurse_3$ & $Nurse_4$ & $Nurse_1$ & $Nurse_2$  \\[1ex] \hline
		$Day_{13}$ & $Nurse_4$ & $Nurse_1$ & $Nurse_2$ & $Nurse_3$ \\[1ex] \hline
		$Day_{14}$ & $Nurse_1$ & $Nurse_3$ & $Nurse_2$ & $Nurse_4$  \\[1ex] \hline
	\end{tabular} \label{NBhData}
\end{minipage}%
\begin{minipage}{0.55\textwidth}
  \centering
\caption{Testing Data} 
\begin{tabular}{ | c | c | c| c| c| c | c |}\hline
 	& $Shift_1$ & $Shift_2$ & $Shift_3$ & $Shift_4$ \\[1ex] \hline
        $Day_{15}$ & $Nurse_3$ & $Nurse_2$ & $Nurse_4$ & $Nurse_1$  \\[1ex]\hline
		$Day_{16}$ & $Nurse_1$ & $Nurse_2$ & $Nurse_4$ & $Nurse_3$  \\[1ex]\hline
		$Day_{17}$ & $Nurse_3$ & $Nurse_4$ & $Nurse_1$ & $Nurse_2$  \\[1ex]\hline
		$Day_{18}$ & $Nurse_3$ & $Nurse_1$ & $Nurse_2$ & $Nurse_4$  \\[1ex]\hline
		$Day_{19}$ & $Nurse_3$ & $Nurse_4$ & $Nurse_2$ & $Nurse_1$  \\[1ex]\hline
		$Day_{20}$ & $Nurse_2$ & $Nurse_1$ & $Nurse_4$ & $Nurse_3$  \\[1ex]\hline
	\end{tabular} \label{NBtestData}
 \end{minipage}%
\end{table}

\noindent\emph{\bf{Predicting the assignment of Shift 4 for row 15:}}\\
\noindent\emph{\bf{Nurse 1:}}
\begin{equation*} 
\begin{split}
P(N_{14} | N_{31}, N_{22}, N_{43}) & = P(N_{31}, N_{22}, N_{43} | N_{14}) . P(N_{14}) \\
 & = P(N_{31} | N_{14}) . P(N_{22} | N_{14}) . P(N_{43} | N_{14}) . P(N_{14}) \\
 & = \frac{1+1}{2+4}  .  \frac{1+1}{2+4}  .  \frac{1+1}{2+4}  . \frac{2}{14} \\
 & = 0.0052
\end{split}
\end{equation*}

\begin{itemize}[leftmargin=*, label=\textbullet]
  \item $P(N_{14}) = \frac{2}{14}$ 
  \inlineitem $P(N_{31} | N_{14}) = \frac{1}{2}$
  \inlineitem $P(N_{22} | N_{14}) = \frac{1}{2}$
  \inlineitem $P(N_{43} | N_{14}) = \frac{1}{2}$
\end{itemize}

\noindent\emph{\bf{Nurse 2:}}
\begin{equation*} 
\begin{split}
P(N_{24} | N_{31}, N_{22}, N_{43}) & = P(N_{31}, N_{22}, N_{43} | N_{24}) . P(N_{24}) \\
 & = P(N_{31} | N_{24}) . P(N_{22} | N_{24}) . P(N_{43} | N_{24}) . P(N_{24}) \\
 & = \frac{1+1}{3+4} . \frac{0+1}{3+4}  . \frac{0+1}{3+4} .  \frac{3}{14} \\
 & = 0.0012 
\end{split}
\end{equation*}

\begin{itemize}[leftmargin=*,label=\textbullet]
  \item $P(N_{24}) = \frac{3}{14}$
  \inlineitem $P(N_{31} | N_{24}) = \frac{1}{3}$
  \inlineitem $P(N_{22} | N_{24}) = \frac{0}{3}$
  \inlineitem $P(N_{43} | N_{24}) = \frac{0}{3}$
\end{itemize}

\noindent\emph{\bf{Nurse 3:}}
\begin{equation*} 
\begin{split}
P(N_{34} | N_{31}, N_{22}, N_{43}) & = P(N_{31}, N_{22}, N_{43} | N_{34}) . P(N_{24}) \\
 & = P(N_{31} | N_{34}) . P(N_{22} | N_{34}) . P(N_{43} | N_{34}) . P(N_{34}) \\
 &= \frac{0+1}{5+4} . \frac{3+1}{5+4} . \frac{2+1}{5+4} .  \frac{5}{14}\\
 & = \bf{0.0058}
\end{split}
\end{equation*}

\begin{itemize}[leftmargin=*,label=\textbullet]
  \item $P(N_{34}) = \frac{5}{14}$
  \inlineitem $P(N_{31} | N_{34}) = \frac{0}{5}$
  \inlineitem $P(N_{22} | N_{34}) = \frac{3}{5}$
  \inlineitem $P(N_{43} | N_{34}) = \frac{2}{5}$
\end{itemize}

\noindent\emph{\bf{Nurse 4:}}
\begin{equation*} 
\begin{split}
P(N_{44} | N_{31}, N_{22}, N_{43}) & = P(N_{31}, N_{22}, N_{43} | N_{44}) . P(N_{44}) \\
 & = P(N_{31} | N_{44}) . P(N_{22} | N_{44}) . P(N_{43} | N_{44}) . P(N_{44}) \\
 &= \frac{1+1}{4+4} . \frac{0+1}{4+4} . \frac{0+1}{4+4} . \frac{4}{14}\\
 & = 0.0011
\end{split}
\end{equation*}

\begin{itemize}[leftmargin=*,label=\textbullet]
  \item $P(N_{44}) = \frac{4}{14}$
  \inlineitem $P(N_{31} | N_{44}) = \frac{1}{4}$
  \inlineitem $P(N_{22} | N_{44}) = \frac{0}{4}$
  \inlineitem $P(N_{43} | N_{44}) = \frac{0}{4}$
\end{itemize}
$\Rightarrow$ \textbf{Nurse 3} will be assigned to shift 4.\\

Table. \ref{NBpredresults} shows the selected class labels with the highest probability for each testing instance. The confusion matrix is given in Figure. \ref{confusionmatrix} where the class labels are numerically encoded as follows; 0: Nurse 1, 1: Nurse 2, 2: Nurse 3, 3: Nurse 4. The accuracy is 66\%, and it is computed according to the positive and negative hits yielded by the ground truth and the predicted results. The accuracy is acceptable given that Naive Bayes is known as instance learning which means the accuracy may be improved with more training data.

\begin{table}[ht]
\centering
\caption{Predicted Results} 
\label{NBpredresults}
	\begin{tabular}{ | c | c | c | c | c | c | c |}
 		\hline
        & Ground Truth & Predicted Results & Positive/Negative Hits \\[1ex] \hline
        $Day_{15}$ & $Nurse_1$ & $Nurse_3$  &  Negative  \\[1ex] \hline
		$Day_{16}$ & $Nurse_3$ & $Nurse_3$  &  Positive  \\[1ex] \hline
		$Day_{17}$ & $Nurse_2$ & $Nurse_2$  &  Positive  \\[1ex] \hline
		$Day_{18}$ & $Nurse_4$ & $Nurse_4$  &  Positive  \\[1ex] \hline
		$Day_{19}$ & $Nurse_1$ & $Nurse_2$  &  Negative  \\[1ex] \hline
		$Day_{20}$ & $Nurse_3$ & $Nurse_3$  &  Positive  \\[1ex] \hline
		Total &  &  &\makecell{Positive = 4\\Negative = 2\\Accuracy = 66\%}  \\[1ex] \hline
	\end{tabular}
\end{table}

\begin{figure}[ht]
\centering
\includegraphics[width=80mm,scale=0.5]{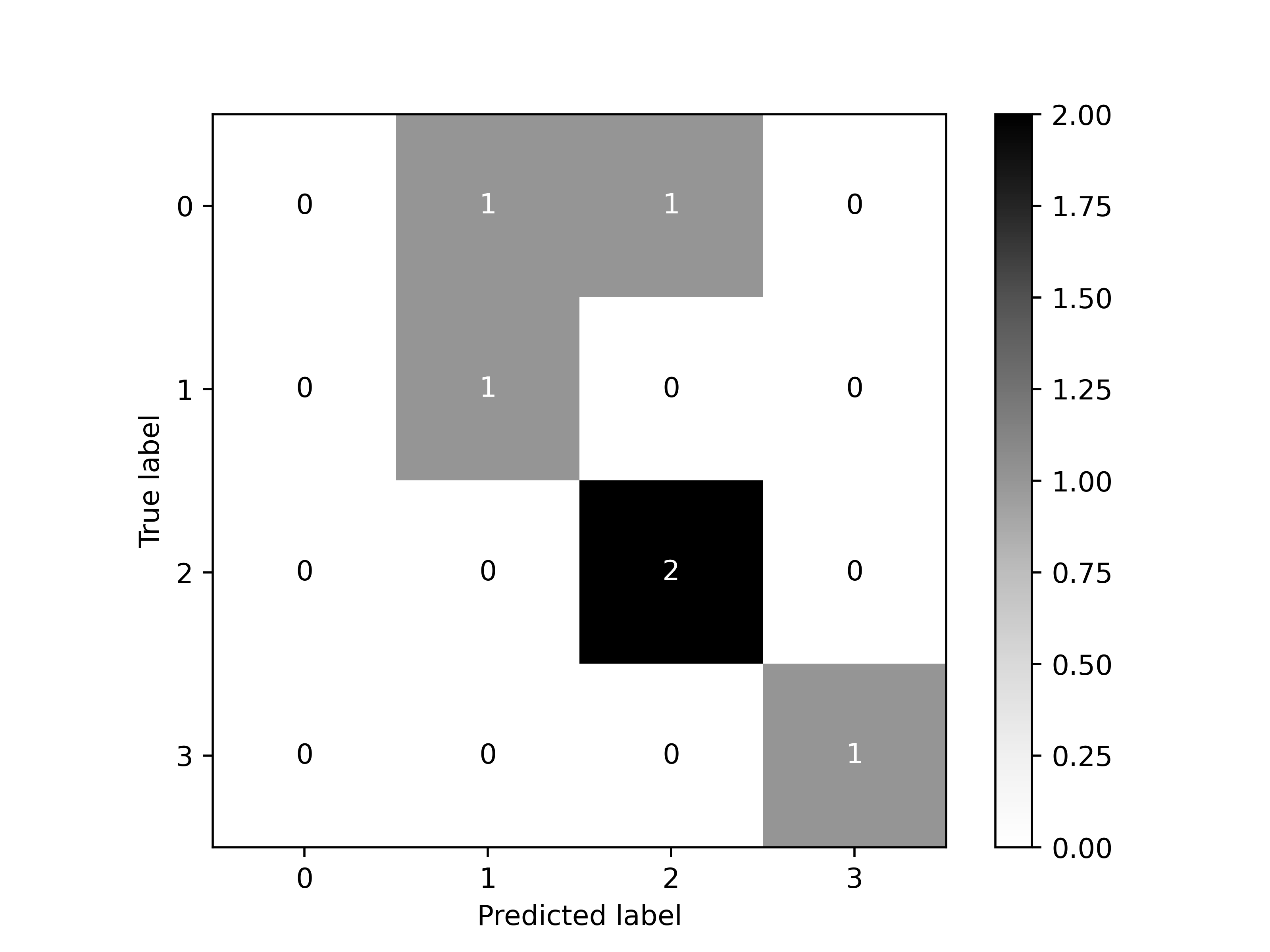}
\caption{Confusion Matrix of Example. \ref{ex2}} \label{confusionmatrix}
\end{figure}

\end{exmp}

\subsection{Bayesian Network (BN)}
A Bayesian Network (BN) can be described as a Directed Acyclic Graph (DAG). The nodes represent random variables and the directed edges are causal relationships between these variables. A probabilistic graphical model provides a compact and intuitive representation of the joint probability distribution across a set of random variables. It simplifies the depiction of causal relationships among these variables. Based on Markov's assumption \cite{K}, only the children nodes are conditioned by their parent nodes. Equation \eqref{BNeq} depicts the calculation of the joint probability distribution in a given BN. Let $X = \{X_{1},...,X_{n}\}$ be a set of variables, $x_{i}$ the values of the variable $X_{i}$, and $Parents(X_{i})$ the values for the parents of $X_{i}$ in the BN.
\begin{equation}
\label{BNeq}
P(x_{1},...,x_{n}) =  \prod_{i=1}^{n} P(x_{i} \ | \ Parents(X_{i}))
\end{equation}

Using a BN to model the relationship between the variables in the NSP historical data is motivated by the fact that the NB method is limited to the distribution of the available schedules. Moreover, the NB model is defined by the product of prior and likelihood and it is a special case of the Bayesian Network since it assumes the independence of variables, however, in the case of NSP, we believe that the variables may be dependent to each other such that assigning specific nurses to particular shifts may have an impact on assigning other nurses to other shifts due to some constraints and objectives. Given that exact inference in BN is NP-hard, we assume that the likelihood function follows a specific distribution and encode its data generating process to simulate new scheduling solutions. Note that the data generating process of our NSP data is: Bernoulli likelihood and Uniform probability.

\subsection{Experimentation}
\subsubsection{Data}
The NSPLib benchmark library \cite{lib} provide various dataset settings, including workload coverage requirements for different number of nurses and different planning horizons, as well as preference costs related to daily shifts. For our experimentation, we use instances with the following setting; 25 nurses, 7 days, and 4 shifts/day and further reduce the number of nurses to 5 for simplicity. Note that the coverage requirements is randomly distributed among nurses to introduce fairness.

\subsubsection{Evaluation metrics and Quality Performance Measures}
The main goal of our implicit solving approach is to automate the generation of scheduling solutions while conserving the properties of the input schedule, in other words, we aim to capture and extract frequent patterns from input schedules, and leverage this knowledge to generate new solutions that inherit the learned patterns. Therefore, we evaluate the quality of the solutions by computing the distance between the input schedule and the generated schedules using the Frobenius Norm \cite{L}. Let $M$ and $N$ be two matrices, $m_{ij}$ and $n_{ij}$ their respective entries, the Frobenius Norm quantifies the element-wise average error between two matrices as depicted in Equation \ref{FNeq}. Regarding the NB method, the quality of the solution is achieved by measuring the model accuracy (i.e. comparing the predicted class labels with the ground truth).

\begin{equation}
\label{FNeq}
\lVert{M - N}\lVert_F = \sqrt{\sum_{i=1}^{x}  \sum_{j=1}^{y} (m_{ij} - n_{ij})^2} 
\end{equation}

\subsection{Results and Discussion}
The experimental results of all our proposed methods are reported in Table~\ref{MLresults}. The Apriori, Two-Phase, and Bayesian Network methods are evaluated using the Frobenius Norm, and Naive Bayes is evaluated based on accuracy. The results reveal low average error in all the methods, while the Naive Bayes accuracy is promising given the limited data.
\begin{table}[ht]
\caption{Experiment results of all methods} 
\label{MLresults}
\centering
\begin{tabular}{| c | c | c | }
    \hline
    Method & Settings & Quality Measure \\ \hline
    Apriori & \makecell{Support: 0.25, Confidence: [0.6 to 0.7] \\Support: 0.25, Confidence: [0.7 to 0.8]\\Support: 0.25, Confidence: [0.8 to 1]} & \makecell{17.94\\20\\17.94} \\ \hline
    Two-Phase & Min-Utility: 120 & 21.40  \\ \hline
    Bayesian Network & \makecell{Probability: Uniform\\Likelihood: Bernoulli} & 20.83  \\ \hline
    Naïve Bayes & \makecell{Training: 70\%\\ Testing: 30\%} & \makecell{Positive hits: 11\\ Negative hits: 5\\ Accuracy: 68\%}  \\ \hline
\end{tabular}
\end{table}

\begin{figure}[ht]
\includegraphics[width=\textwidth]{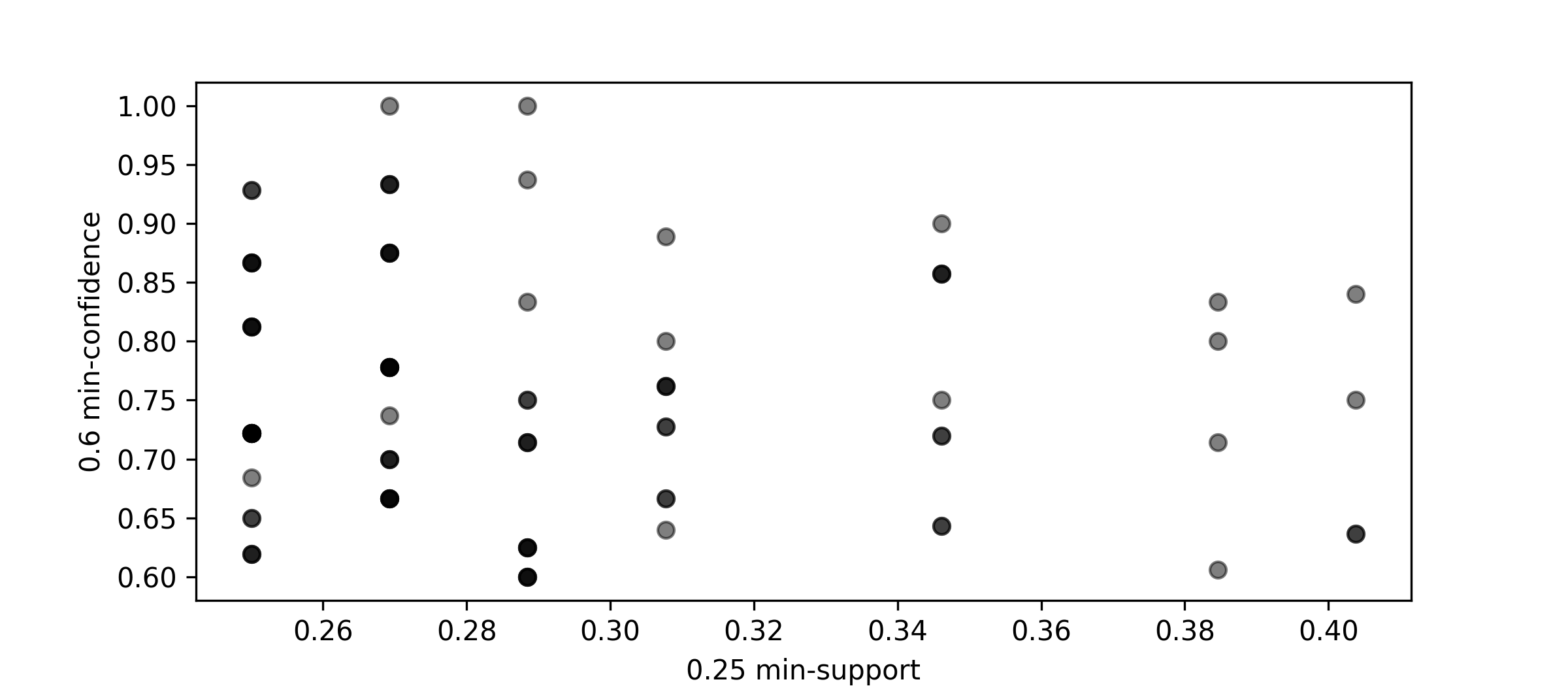}
\caption{Rules Distribution} \label{0.25s0.6c}
\end{figure}

\begin{figure}[ht]
\includegraphics[width=\textwidth]{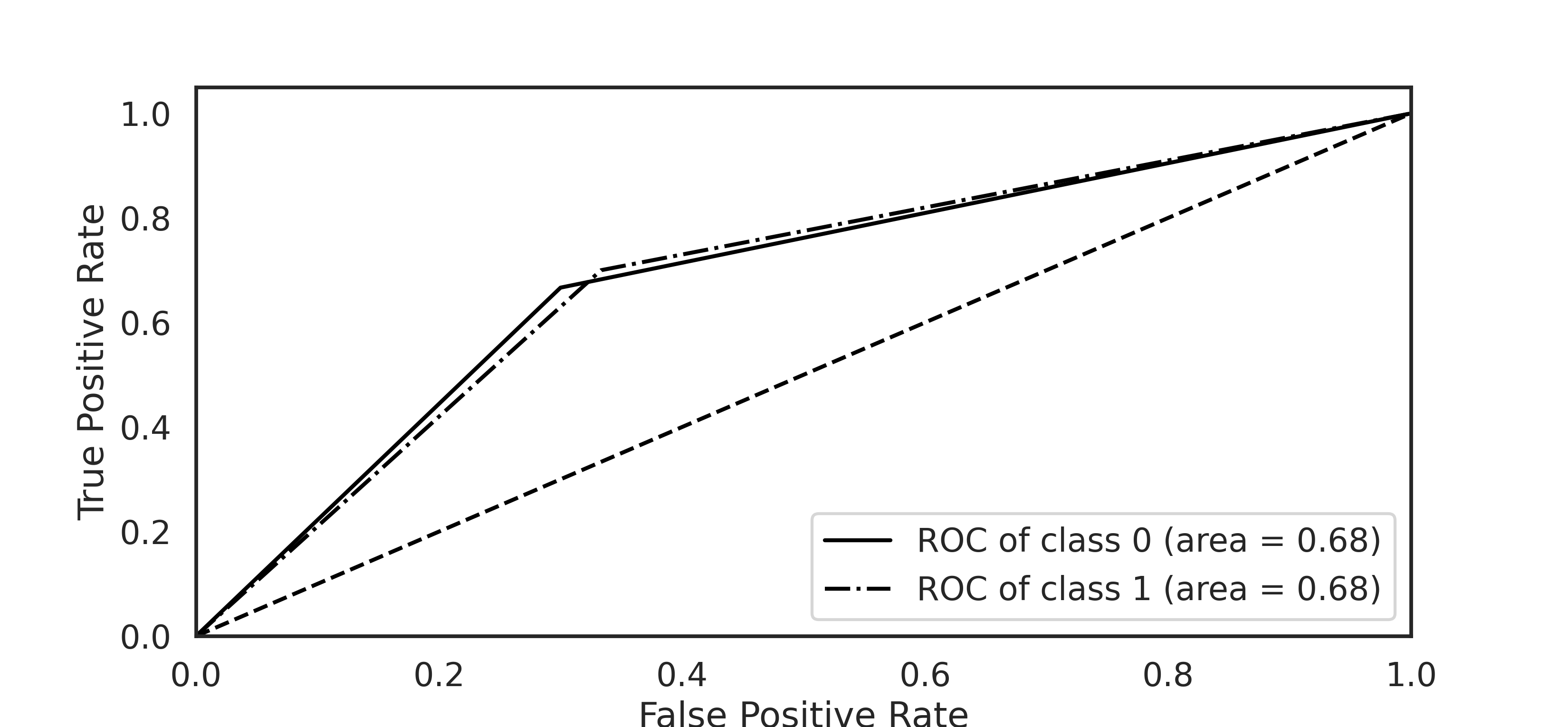}
\caption{ROC curve for class 0 and 1} \label{roc}
\end{figure}

\begin{figure}[ht]
\includegraphics[width=\textwidth]{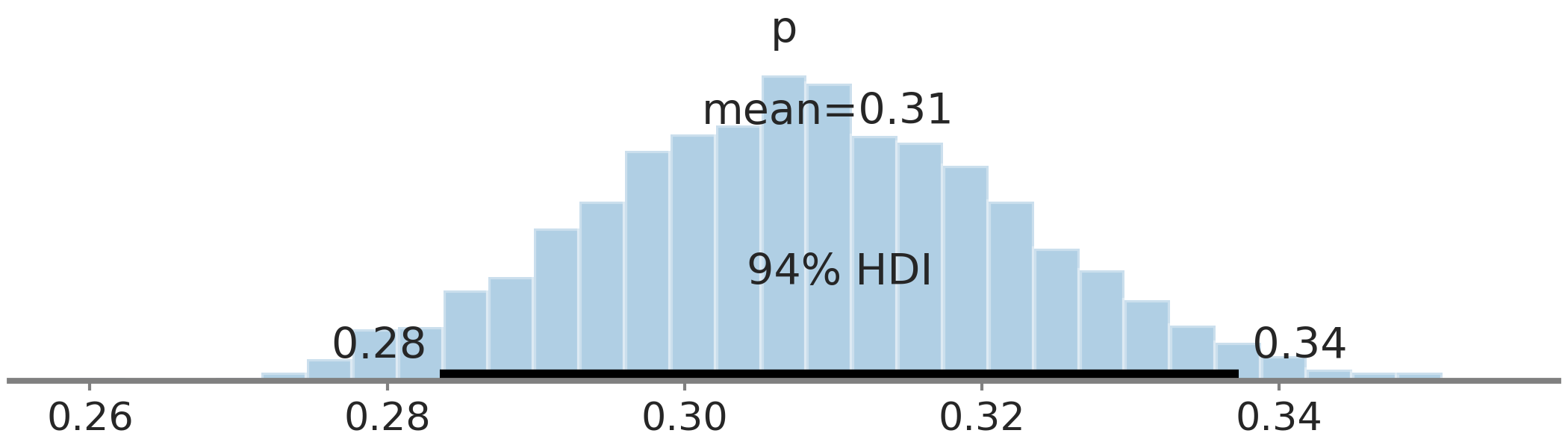}
\caption{Highest Density Interval (HDI) of Posterior Distribution}\label{fig5}
\end{figure}

The FN values represents the average error element-wise between the input and the generated schedules. Ideally, a value closer to 0 indicates that the obtained schedules closely match the original ones. However, the results of our methods, as reported in Table \ref{MLresults}, are acceptable because we do not explicitly define any preferences and use limited data. We plan to conduct a rigorous external validation in the future by consulting with hospital scheduling experts. The schedule generation using our methods is an iterative process that relies on specific parameters. We expect to achieve better results by tuning these measures for Apriori and Two-Phase, or by increasing sampling from the Bayesian Network. Therefore, the next phase of research may focus on learning how to find optimal or sub-optimal solutions without iterations by investigating the correlation between the quality of the generated solutions and the number of iterations. The ROC curve presented in Fig.\ref{roc} outlines the trade-off between the sensitivity and the specificity of our NB model for the binary class labels. Ideally the ROC curve should be closer to the true positive rate axis indicating that the higher probabilities are assigned to the correct class label. The accuracy of our NB model is 68\%, however, NB is known as instance based learning which means that the accuracy of the model is dependent on the training set; the better quality of data the better accuracy. Fig.\ref{fig5} shows the posterior distribution from Markov Chain Monte Carlo (MCMC) sampling, representing the convergence of the prior probabilities after seeing more evidence from the data. The plot reveals that 94\% of the probability is within the 0.28 to 0.34 range, showing that likelihood values within this interval yield the highest accuracy.


In addition to comparing the proposed ML methods individually, we also conduct an experiment against COUNT-OR \cite{kumar}. The quality of the generated schedules is assessed using the Frobenius Norm, and the result is as follows (in this experiment, we exclude the NB method due to incomparable quality metric); Apriori: 5.83, Two-Phase: 6.63, Bayesian Network: 6.70, COUNT-OR: 6.78. Overall, our methods performed slightly better because ML algorithms are known for capturing frequent patterns from the data, which lead to generating instances similar to historical data. Additionally, COUNT-OR explicitly learns the constraints of input schedules against predefined quantities of interest (obtained using tensor operations). Consequently, if the input schedules do not satisfy the predefined quantities, COUNT-OR may fail to produce a solution. In contrast, our methods implicitly learn the constraints and preferences, disregarding any inconsistencies in the data because they are designed to extract the highly interesting frequent patterns. Although our methods may not always provide an optimal solution, they guarantee a solution based on the given data. Furthermore, since the constraints and objectives are represented implicitly through the learned data patterns, our proposed approach may come with uncertainty concerning which constraints are satisfied and which objectives are optimized, and therefore, may not guarantee an optimal solution. To tackle the uncertainty limitation related to out ML implicit approach, we propose an alternative approach using the CSP framework.

\section{Explicit Solving of the NSP}
\label{explicitapproach}

The CSP  is a powerful framework for modeling and solving real-world constraint satisfaction problems such as Map Coloring, N-Queen, TSP, Knapsack, etc. Modeling such problems using the CSP framework consists of formulating it in terms of variables $X_{i} = \{x_{1},...,x_{n}\}$, defined on a set of non-empty domains of possible values $D_{i} = \{dom(x_{1}),...,dom(x_{n})\}$, and a set of constraints $C = \{c_{1},...,c_{k}\}$ restricting variables' assignment combinations. The goal of solving a CSP is to find a consistent assignment of values to the variables from their domains such that all the constraints are satisfied. The WCSP is an extension of the CSP which considers violation costs related to soft constraints or weights associated with the domain values. In addition to finding a solution that satisfies all the constraints, the target of a WCSP is to optimize the solution's cost. Solving CSPs and WCSPs usually involves exact and approximate methods (e.g. backtracking, local search, SLS, etc) that work based on global and local search respectively, to elicit candidate solutions. Exact methods guarantee the optimal solution but come with a heavy running time cost while approximate methods often trade the quality of the solution for a better running time.  Approximate methods iterate through the search space to improve the most recent solution while exact methods explore the entire search space to find the optimal one. Given the fact that the exact solving techniques require exponential running time to find the optimal solution, researchers usually rely on various constraint propagation techniques (e.g. NC, AC, GAC, etc), and Variables/values ordering heuristics in the scope of CSPs and WCSPs to minimize the search space which consequently optimizes the running time. In this context, we rely on the WCSP framework to model the NSP in terms of variables, domains, and constraints, and solve it using explicit methods. For the solving task, we propose a variant of the B\&B and SLS methods as exact and approximate solvers for the WCSP. Furthermore, we also optimize the search domain relying on constraints propagation techniques to minimize the execution time related to our proposed methods, particularly B\&B, since it may require exploration of the entire search space to find the optimal solution in the worst-case. In addition to our proposed methods, we conduct further experiments against approximate methods (Whale Optimization Algorithm (WOA) and Genetic algorithm (GA)) \cite{10.1007/978-3-031-34020-8_5,icores24} to asses the efficiency of our methods in solving the NSP in terms of quality of solution and running time.

\subsection{WCSP Problem Formulation}
\label{WCSPFormulation}
The WCSP is defined by the tuple $(X,D,C,K)$, where $X$, $D$, and $C$ represent the variables, domains, and constraints, respectively. $K$ represent the largest numerical value for the cost decision variable $c_{ij}.$\\

\noindent  \emph{\bf Variables:} \text{$X = \{X_{1},...,X_{n}$\} is the set of nurses.}

\noindent  \emph{\bf Domain:} \text{D = the set of all possible shift patterns.}

\noindent  \emph{\bf Constraints:} \text{$C = \{const_{1},...,const_{4}$\}  is the set of NSP constraints}\\

\noindent To model the NSP constraints, we rely on function $A(i,j,k,s)$ as described below. 
\begin{equation*}
 A(i,j,k,s) =
    \begin{cases}
      1,& \text{if nurse i $(X_i)$ is assigned shift pattern j, and j covers shift s on day k}\\
      0,& \text{Otherwise}\\
    \end{cases}      
\end{equation*}

\noindent Our WCSP problem formulation consist of global and unary constraints: $const_1$ is a global constraint (involving all the variables) and constraints $const_2, const_3, const_4$ are unary constraints (involving individual variables). The following parameters and indices are used to elicit the constraints.\\

\noindent  \emph{\bf Parameters and Indices}
\begin{equation*}
\begin{cases}
\text{$n =$ Number of nurses}\\
\text{$m =$ Number of possible shift patterns}\\
\text{$c_{ij} = $ Cost of assigning nurse i the shift pattern j}\\
\text{$q_{sk} = $ Minimum nurses needed for shift $s$ in day $k$} \\
\text{$p_{sk} = $ Maximum number of nurses required for shift $s$ in day $k$} \\
\text{$h_{i} = $ Maximum number of shifts for nurse i during the schedule}\\
\text{$y = $ Maximum number of consecutive shifts (night shift followed by a morning shift)}\\
\text{$b_{i} = $ Maximum number of night shifts for nurse i during the schedule}\\
\text{$i = \{1,..,n\}$ is the nurse index}\\
\text{$j = \{1,...,m\}$ is the index of the weekly shift pattern}\\
\text{$k = \{1,..,7\}$ is the day index} \\
\text{$s = \{1,..,3\}$ is the shift index within a given day} \\
\text{$z = \{1,..,21\}$ is the index of shifts in a given shift pattern} \\
\end{cases}      
\end{equation*}\\

\noindent  \emph{\bf Hard Constraints:}
\begin{enumerate}
\item Minimum and Maximum number of nurses per shift

\begin{equation}
const_{1}: p_{sk} \leq \sum_{i=1}^{n} A(X_{i},a_{j_{i}},k,s)  \geq \ q_{sk} \ \ ,\forall k, \forall s, a_{j_{i}} \in D 
\end{equation}

\item Maximum number of shifts for a given nurse during the schedule

\begin{equation}
const_{2}: \sum_{k=1}^{7} \sum_{s=1}^{3} A(X_{i},a_{j_{i}},k,s)  \leq \ h_{i} \ \ ,\forall X_{i}, a_{j_{i}} \in D 
\end{equation}

\item Maximum number of consecutive shifts (night shifts followed by morning shifts)

\begin{equation}
const_{3}: \sum_{k=1}^{6} A(X_{i},a_{j_{i}},k,3) + A(X_{i},a_{j_{i}},k+1,1) \leq y \ \ ,\forall X_{i},  a_{j_{i}} \in D 
\end{equation}

\item Maximum number of night shifts

\begin{equation}
const_{4}: \sum_{k=1}^{7} A(X_{i},a_{j_{i}},k,3)  \leq \ b_{i} \ \ ,\forall X_{i}, a_{j_{i}} \in D 
\end{equation}

\end{enumerate}

\noindent  \emph{\bf Soft Constraint:}

\begin{equation}
f_i: a_{j_{i}} \in D \rightarrow c_{ij_{i}}
\end{equation}

\noindent  \emph{\bf Objective:} hospital costs to minimize

\begin{equation}
 Minimize( \sum_{i=1}^{n} c_{ij_{i}} ) \  a_{j_{i}} \in D  
\end{equation}

\subsection{Branch \& Bound (B\&B)}
B\&B uses the Depth First Search (DFS) strategy to explore all candidate solutions. This process involves generating sub-branches and applying the pruning concept using the Lower Bound (LB) and Upper Bound (UB) parameters. The UB and LB are used in the pruning process to verify in advance if the exploration of a branch would result in getting a better solution or not before actually traversing the branch, preventing the exploration of branches that do not lead to an optimal solution. In our minimization problem (minimizing hospital's costs), the LB overestimates the best possible solution. The UB represents the best solution found so far and it is updated whenever a new solution with a better objective value is discovered. Initially, the LB value is set based on the least significant predictable weighted solution as suggested in \cite{Stefan2020ABAf,BruscoMichaelJ2005BAiC,HaouariM2005Stgm}. During search, the LB is estimated by computing the weight of the traversed sub-branch plus the minimum values' weights that could possibly be assigned to the remaining nodes in the corresponding tree path, obtained from the cost matrix $c_{ij}$. The LB parameter contribute to the pruning process during the execution of B\&B such that; if the estimated LB becomes greater than or equal to UB, then the algorithm will backtrack without continuing the exploration of the recent decision because the solution will not be better than the one already found. Otherwise, if LB is less than UB, then the algorithm keep exploring the recent decision because it may lead to a better solution. This approach may be viewed as a form of Forward Checking to prevent future failures, significantly reducing the search space and consequently minimizing the running time. The pseudo-algorithm of our minimization B\&B variant is presented in Algorithm~\ref{BnB}.

\begin{algorithm}[ht]
\SetAlgoLined
$Variables = \{$1,...,n$\}$\\
$Domain = \{$1,...,m$\}$\\
$OptSol = \{\emptyset\}$\\
$TmpSol = \{\emptyset\}$\\
$UB = \infty$ \\
$LB = \infty$\\
$Branch\&Bound (TmpSol, LB, Domain)$\\
\hspace{0.2 in}${\bf if}\ length(TmpSol) = length(Variables)$\\
\hspace{0.33 in}${\bf if}\ LB \leq UB \ {\bf and} \  GlobalConst(TmpSol) $\\
\hspace{0.55 in}$UB = LB$\\
\hspace{0.55 in}$OptSol = \{\emptyset\}$\\
\hspace{0.55 in}${\bf for} \ x \ {\bf in} \ TmpSol$\\
\hspace{0.7 in}$OptSol.Add(x)$\\
\hspace{0.33 in}{\bf Return} True\\
\hspace{0.2 in}${\bf if} \ LB \geq UB$\\
\hspace{0.33 in}{\bf Return} True\\
\hspace{0.2 in}${\bf for} \ value \ {\bf in} \ Domain$\\
\hspace{0.33 in}$TmpSol.Add(value)$\\
\hspace{0.33 in}$LB\ = ComputeLB(TmpSol)$\\
\hspace{0.33 in}$Branch\&Bound(TmpSol, LB, Domain)$\\
\hspace{0.33 in}$TmpSol.Remove(value)$\\
{\bf Return}  $OptSol, UB$
 \caption{The Branch \& Bound Pseudo-Algorithm \cite{NSPCSP}}
\label{BnB}
\end{algorithm}

\begin{figure}[htb]
\begin{center}
\begin{tikzpicture}
\node[draw, circle,minimum size=0.7cm, label= right:{UB=$\infty$}] (A) at (-6,8.5) {root};
\node[draw, circle,minimum size=0.7cm] (B) at (-8,7) {1};
\node[draw, circle,minimum size=0.7cm, label= right: X1] (C) at (-4,7) {m};
\path (B) -- node[auto=false]{\ldots} (C);
\node[draw, circle,minimum size=0.7cm] (D) at (-9,5.5) {1};
\node[draw, circle,minimum size=0.7cm] (E) at (-7,5.5) {m};
\path (D) -- node[auto=false]{\ldots} (E);
\node[draw, circle,minimum size=0.7cm] (F) at (-5,5.5) {1};
\node[draw, circle,minimum size=0.7cm, label= right: X2] (G) at (-3,5.5) {m};
\path (F) -- node[auto=false]{\ldots} (G);
\path (E) -- node[auto=false]{\ldots} (F);
\node[draw, circle,minimum size=0.7cm] (H) at (-10,4) {1};
\node[draw, circle,minimum size=0.7cm] (I) at (-8,4) {m};
\path (H) -- node[auto=false]{\ldots} (I);
\node[draw, circle,minimum size=0.7cm] (J) at (-4,4) {1};
\node[draw, circle,minimum size=0.7cm, label= right: X3] (K) at (-2,4) {m};
\path (J) -- node[auto=false]{\ldots} (K);
\path (I) -- node[auto=false]{\ldots\ldots\ldots\ldots} (J);
\node[draw, circle,minimum size=0.7cm] (L) at (-11,2.5) {1};
\node[draw, circle,minimum size=0.7cm] (M) at (-9,2.5) {m};
\path (L) -- node[auto=false]{\ldots} (M);
\node[draw, circle,minimum size=0.7cm] (N) at (-3,2.5) {1};
\node[draw, circle,minimum size=0.7cm, label= right: X4] (O) at (-1,2.5) {m};
\path (N) -- node[auto=false]{\ldots} (O);
\path (M) -- node[auto=false]{\ldots\ldots\ldots\ldots\ldots\ldots\ldots\ldots} (N);
\node[] (P) at (-11.75,9) {};
\node[] (Q) at (-11.75,2) {};
\draw (P)  -- node [sloped, anchor=center, above, text width=4cm] {Path is Subject to $c_{2}$\\LB is Subject to $obj$ } (Q);

\draw (A)  -- node [sloped, anchor=center, above, text width=2.5cm] {} (B);
\draw (A)  --  node [sloped, anchor=center, above, text width=2.5cm] {} (C);
\draw (B)  -- node [sloped, anchor=center, above, text width=2.5cm] {} (D);
\draw (B)  --  node [sloped, anchor=center, above, text width=2.5cm] {} (E);
\draw (C)  -- node [sloped, anchor=center, above, text width=2.5cm] {} (F);
\draw (C)  --  node [sloped, anchor=center, above, text width=2.5cm] {} (G);
\draw (D)  -- node [sloped, anchor=center, above, text width=2.5cm] {} (H);
\draw (D)  --  node [sloped, anchor=center, above, text width=2.5cm] {} (I);
\draw (G)  -- node [sloped, anchor=center, above, text width=2.5cm] {} (J);
\draw (G)  --  node [sloped, anchor=center, above, text width=2.5cm] {} (K);
\draw (H)  -- node [sloped, anchor=center, above, text width=2.5cm] {} (L);
\draw (H)  --  node [sloped, anchor=center, above, text width=2.5cm] {} (M);
\draw (K)  -- node [sloped, anchor=center, above, text width=2.5cm] {} (N);
\draw (K)  --  node [sloped, anchor=center, above, text width=2.5cm] {} (O);

\end{tikzpicture}
\caption{Branch \& Bound Solving Tree} \label{BnBsolTree}
\end{center}
\end{figure}
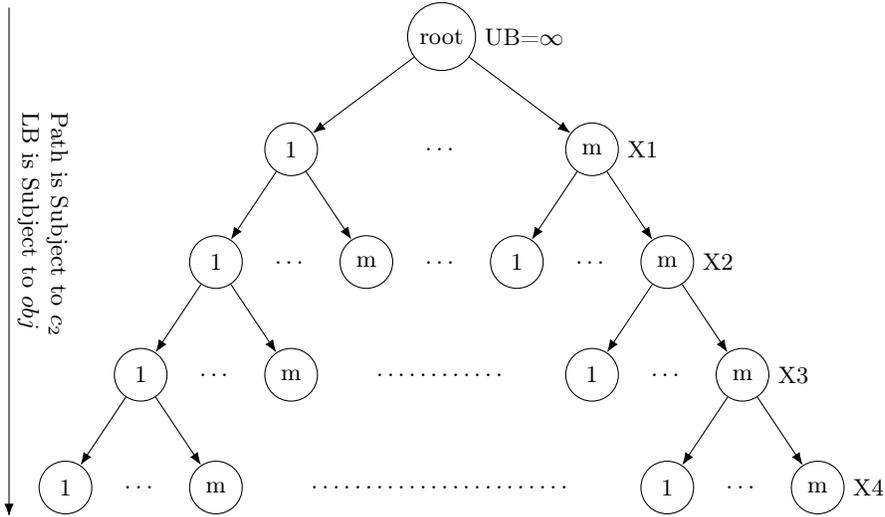

Figure~\ref{BnBsolTree} visualizes the solving tree of our B\&B variant. Each stage of the tree corresponds to a random variable $X_{i}$, with the tree depth representing the total number of variables. The nodes represent the values assigned to each variable, and the width denotes the domain size. As previously mentioned, B\&B employs a DFS search to explore all candidate solutions, thus, we rely on a recursive search procedure to traverse all tree nodes. An optimal solution is formed by a set of nodes representing a path from the root node to a leaf node, such that the combination of values assigned to the variables is consistent with all constraints and maximizes the objective function.

\subsection{Constraint Propagation (CP)}
\label{CP}
CP relies on local consistency techniques \cite{dechter2003constraint,DBLP:conf/appinf/Mouhoub03} to eliminate some ``locally'' inconsistent domain values and consequently reduce the search space scope. Enforcing CP may lead to two outcomes; either resulting in an empty domain, which indicates the inconsistency of the WCSP, or leaving multiple values in some domains, which requires the use of B\&B to solve the problem and find the optimal solution. Various techniques are employed to enforce local consistency, including Arc Consistency, Path Consistency, etc. Considering the the types of constraints in our WCSP (unary and global constraints), we utilize two local consistency algorithms: Node Consistency (NC) and Generalized Arc Consistency (GAC). In addition to local consistency, we rely on variables ordering heuristics to help find the optimal solution faster following the ``fail first principle'', prioritizing the variables that most likely lead to a dead-end \cite{DBLP:journals/apin/YongM18,mouhoub2011heuristic}, and ordering the domain values according to their weights. Node consistency is used to reduce variables' domain size by eliminating all the values that violate the unary constraint. A WCSP is node consistent if all its variables are node consistent \cite{Larrosa} (i.e. every value in every variable' domain satisfies the unary constraint(s)). In addition to NC,  we also apply a GAC algorithm \cite{BessiereR97,Regin96} to tackle the global constraint $c_2$. GAC is an extension of Arc Consistency (AC), and it is recognized as an effective local consistency technique for solving CSPs, particularly when combined with backtrack search. GAC is enforced through k-ary constraints (where k is strictly greater than 2) to eliminate every value that does not have a consistent combination of values from the rest of the variables in the scope of the k-ary constraints. Therefore, any values in the domains of variables that do not have at least one feasible combination of values across the remaining variables should be eliminated. A WCSP is generalized arc consistency when every value within each variable's domain can be matched with at least one combination of values from the other variables involved in scope of the global constraint.

\begin{exmp}\label{ex1}
Let us consider a WCSP with two variables/nurses $N = \{N_1, N_2\}$ defined on the domain $D = \{1001, 0100, 0110\}$ representing the coverage of four shifts during a single day where every value is associated with a preference cost/weight, $Costs = \{1001: 2, 0100: 1, 0110: 4\}$. The target is to assign the nurses with feasible shift patterns such that the solution satisfies the minimum shift coverage requirements while maximizing the total preference cost. Branch and Bound (B\&B) will be used as an exact solver for this problem. \\
{\bf Shifts minimum requirements:} $\{shift_{1}: 1, shift_{2}: 1, shift_{3}: 1, shift_{4}: 1\}$\\ 
\noindent  \emph{\bf Global constraint:} The solution must satisfy the minimum shift coverage requirements.\\
\emph{\bf Unary constraint:} Every nurse must at least work two shifts in a single day.\\
\noindent  \emph{\bf Objective function:} $Maximize(Cost(N_{1}, N_{2})) $

\begin{figure}[ht]
\begin{center}
\begin{tikzpicture}[-,auto,node distance=3cm,main node/.style={draw,font=\bfseries}]

    \draw[dashed] (-1,1.75) ellipse [x radius=0.8, y radius = 2] node[xshift=0cm, yshift=2.5cm] {$N_{1}$};
    \draw[dashed] (3,1.75) ellipse [x radius=0.8, y radius = 2] node[xshift=0cm, yshift=2.5cm] {$N_{2}$};

      \node[draw, rectangle,label={above, text width=0.2cm: $2$}] (A) at (-1,3) {$1001$};
      \node[draw, rectangle,label={above, text width=0.2cm: $1$}] (B) at (-1,1.75) {$0100$};
      \node[draw, rectangle,label={above, text width=0.2cm: $4$}] (C) at (-1,0.5) {$0110$};
      
      \node[draw, rectangle,label={above, text width=0.2cm: $2$}] (D) at (3,3) {$1001$};
      \node[draw, rectangle,label={above, text width=0.2cm: $1$}] (E) at (3,1.75) {$0100$};
      \node[draw, rectangle,label={above, text width=0.2cm: $4$}] (F) at (3,0.5) {$0110$};

      \draw (A) -- node  {} (D) ;
      \draw (A) -- node  {} (E) ;
      \draw (A) -- node  {} (F) ;
      
      \draw (B) -- node  {}(D)  ;
      \draw (B) -- node  {} (E) ;
      \draw (B) -- node  {} (F) ;

      \draw (D) -- node {} (C) ;
      \draw (E) -- node {} (C) ;
      \draw (F) -- node {} (C) ;
     
\end{tikzpicture}
\end{center}
\caption{Weighted CSP of Example~\ref{ex1}}
\end{figure}
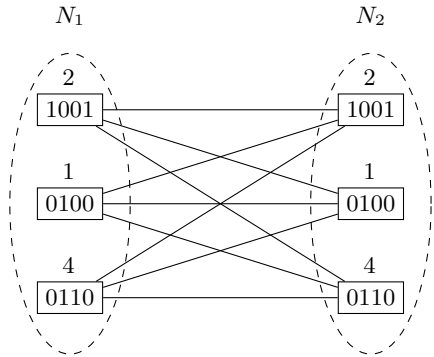
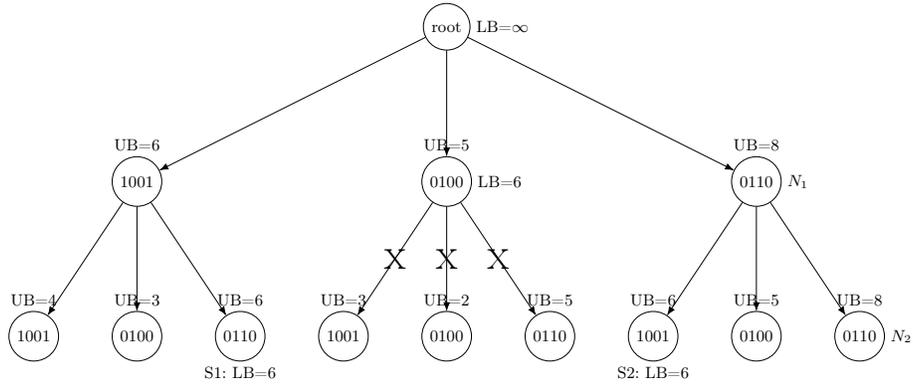
\begin{figure}[ht]
\centering
\resizebox{\linewidth}{!}{
\begin{tikzpicture}
\node[draw, circle,minimum size=0.7cm, label= right:{LB=$\infty$}] (A) at (0,0) {root};

\node[draw, circle,minimum size=0.7cm, label=above: {UB=6}] (B) at (-6,-3) {$1001$};
\node[draw, circle,minimum size=0.7cm, label=above: {UB=5}, label=right: {LB=6}] (C) at (0,-3) {$0100$};
\node[draw, circle,minimum size=0.7cm, label= right: $N_1$, label=above: {UB=8}] (D) at (6,-3) {$0110$};

\node[draw, circle,minimum size=0.7cm, label=above: {UB=4}] (E) at (-8,-6) {$1001$};
\node[draw, circle,minimum size=0.7cm, label=above: {UB=3}] (F) at (-6,-6) {$0100$};
\node[draw, circle,minimum size=0.7cm, label= below: {S1: LB=6}, label=above: {UB=6}] (G) at (-4,-6) {$0110$};

\node[draw, circle,minimum size=0.7cm, label=above: {UB=3}] (H) at (-2,-6) {$1001$};
\node[draw, circle,minimum size=0.7cm, label=above: {UB=2}] (I) at (0,-6) {$0100$};
\node[draw, circle,minimum size=0.7cm, label=above: {UB=5}] (J) at (2,-6) {$0110$};

\node[draw, circle,minimum size=0.7cm, label= below: {S2: LB=6}, label=above: {UB=6}] (K) at (4,-6) {$1001$};
\node[draw, circle,minimum size=0.7cm, label=above: {UB=5}] (L) at (6,-6) {$0100$};
\node[draw, circle,minimum size=0.7cm, label= right: $N_2$, label=above: {UB=8}] (M) at (8,-6) {$0110$};

\node[label=above: {}] (N) at (1,-4.5) {\Huge x};
\node[label=above: {}] (O) at (0,-4.5) {\Huge x};
\node[label=above: {}] (P) at (-1,-4.5) {\Huge x};

\draw (A)  -- node [sloped, anchor=center, above, text width=2.5cm] {} (B);
\draw (A)  -- node [sloped, anchor=center, above, text width=2.5cm] {} (C);
\draw (A)  -- node [sloped, anchor=center, above, text width=2.5cm] {} (D);

\draw (B)  -- node [sloped, anchor=center, above, text width=2.5cm] {} (E);
\draw (B)  -- node [sloped, anchor=center, above, text width=2.5cm] {} (F);
\draw (B)  -- node [sloped, anchor=center, above, text width=2.5cm] {} (G);

\draw (C)  -- node [sloped, anchor=center, above, text width=2.5cm] {} (H);
\draw (C)  -- node [sloped, anchor=center, above, text width=2.5cm] {} (I);
\draw (C)  -- node [sloped, anchor=center, above, text width=2.5cm] {} (J);

\draw (D)  -- node [sloped, anchor=center, above, text width=2.5cm] {} (K);
\draw (D)  -- node [sloped, anchor=center, above, text width=2.5cm] {} (L);
\draw (D)  -- node [sloped, anchor=center, above, text width=2.5cm] {} (M);

\end{tikzpicture}}
\caption{B\&B Solving Tree of Example~\ref{ex1} without Constraint Propagation} \label{BnBEx1}
\end{figure}

The Branch and Bound solving tree is depicted in Figure. \ref{BnBEx1}. After finding the first solution ($S1 = [1001, 0110], \ Cost(S1) = 6$), B\&B pruned the sub-branches of node $0100$ because exploring these branches will not lead to a better solution given that the estimated UB at this node is lower than the LB of the initial solution S1. The rest of the execution resulted in a second feasible solution with an equal LB ($S2 = [0110, 1001], \ \ Cost(S2) = 6$). After enforcing constraint propagation, precisely Node Consistency and Generalized Arc Consistency, the value $``0100"$ was removed from the domain since it violates both the Unary constraint and the global constraint. The new Branch \& Bound solving tree is given in Figure. \ref{BnBEx1CP}.

\begin{figure}[htbp]
\centering
\resizebox{\linewidth}{!}{
\begin{tikzpicture}
\node[draw, circle,minimum size=0.7cm, label= right:{LB=$\infty$}] (A) at (0,0) {root};

\node[draw, circle,minimum size=0.7cm, label=above: {UB=6}] (B) at (-6,-3) {$1001$};

\node[draw, circle,minimum size=0.7cm, label= right: $N_1$, label=above: {UB=8}] (C) at (6,-3) {$0110$};

\node[draw, circle,minimum size=0.7cm, label=above: {UB=4}] (E) at (-8,-6) {$1001$};

\node[draw, circle,minimum size=0.7cm, label= below: {S1: LB=6}, label=above: {UB=6}] (F) at (-4,-6) {$0110$};

\node[draw, circle,minimum size=0.7cm, label= below: {S2: LB=6}, label=above: {UB=6}] (G) at (4,-6) {$1001$};

\node[draw, circle,minimum size=0.7cm, label= right: $N_2$, label=above: {UB=8}] (H) at (8,-6) {$0110$};

\draw (A)  -- node [sloped, anchor=center, above, text width=2.5cm] {} (B);
\draw (A)  -- node [sloped, anchor=center, above, text width=2.5cm] {} (C);
\draw (B)  -- node [sloped, anchor=center, above, text width=2.5cm] {} (E);
\draw (B)  -- node [sloped, anchor=center, above, text width=2.5cm] {} (F);
\draw (C)  -- node [sloped, anchor=center, above, text width=2.5cm] {} (G);
\draw (C)  -- node [sloped, anchor=center, above, text width=2.5cm] {} (H);

\end{tikzpicture}}
\caption{B\&B Solving Tree of Example~\ref{ex1} with Constraint Propagation} \label{BnBEx1CP}
\end{figure}
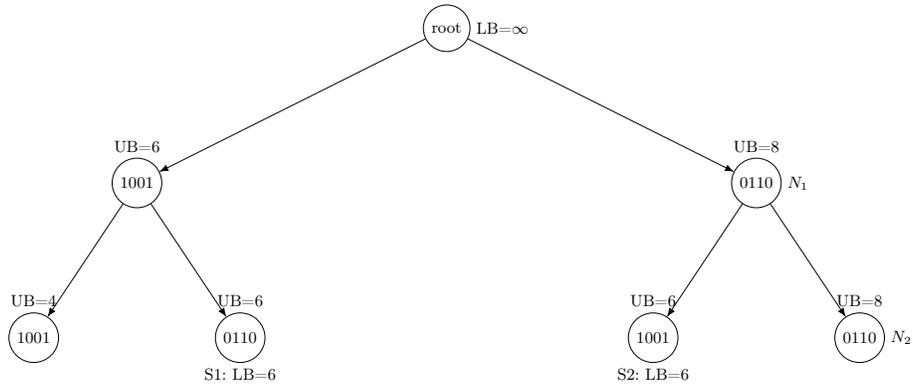

\end{exmp}

\subsection{Stochastic Local Search (SLS)} \label{SLS-basedApproach}
While exact methods guarantee the optimal solution in solving the NSP, they come with exponential time complexity. As a result, researchers often explore strategies to address the exponential time cost associated with exact methods, or rely on approximate methods that trade the quality of the solution for a better running time. We use the NSP formulation presented in \cite{NSPOLA} and further transform it into a WCSP model (see section \ref{WCSPFormulation}), then we propose three SLS-based methods to solve it. In this context, we propose three SLS variants as shown in Figure \ref{SLSVariants}. These variants works by identifying an initial solution, and then tuning it further to enhance the solution's quality while ensuring it remains feasible by systematically selecting an alternative value for each variable from its domain such that the new value would improve the quality of the solution. The difference between the three SLS variants is how the initial solution is obtained (as illustrated in Figures \ref{RandonSearch} and \ref{DFSsearch}). The first variant (we call SLS) uses a random search to obtain the initial solution. The second variant (we call DFS+SLS) returns the first feasible solution through a DFS search. And the third variant (we call DFS + NC + GAC + SLS) uses CP as a pre-processing step on top of DFS to optimize the search space, by removing the domain values that do not contribute to any feasible solution.

\begin{figure}[ht]
\begin{minipage}{0.45\textwidth}
\centering
\resizebox{\linewidth}{!}{%
\begin{tikzpicture}
\node[draw,minimum size=0.7cm, label= right:{}] (A) at (0,0) {Domain = \{1, ,2, 3, ...,m\}};
\node[draw, circle,minimum size=0.7cm, label=above: {}] (E) at (-3,-3) {V1};
\node[draw, circle,minimum size=0.7cm, label= below: {}, label=above: {}] (F) at (-1.5,-3) {V2};
\node[draw, circle,minimum size=0.7cm, label= below: {}, label=above: {}] (I) at (0,-3) {V3};
\node[minimum size=0.7cm, label= below: {}, label=above: {}] (G) at (1.5,-3) {.......};
\node[draw, circle,minimum size=0.7cm, label= right: , label=above: {}] (H) at (3,-3) {Vn};

\draw (A)  -- node [sloped, anchor=center, above, text width=2.5cm] {} (E);
\draw (A)  -- node [sloped, anchor=center, above, text width=2.5cm] {} (F);
\draw (A)  -- node [sloped, anchor=center, above, text width=2.5cm] {} (I);
\draw (A)  -- node [sloped, anchor=center, above, text width=2.5cm] {} (H);
\end{tikzpicture}}
\caption{Random Assignment of Values} \label{RandonSearch}
\end{minipage} \hfill
\begin{minipage}{0.45\textwidth}
\centering
\resizebox{\linewidth}{!}{%
\begin{tikzpicture}
\node[draw, circle,minimum size=0.7cm, label= right:{}] (A) at (0,0) {0};
\node[draw, circle,minimum size=0.7cm, label=above: {}] (B) at (-2,-1) {1};
\node[draw, circle,minimum size=0.7cm, label= right: , label=above: {}] (C) at (2,-1) {4};
\node[draw, circle,minimum size=0.7cm, label=above: {}] (E) at (-3,-3) {2};
\node[draw, circle,minimum size=0.7cm, label= below: {}, label=above: {}] (F) at (-1,-3) {3};

\node[draw, circle,minimum size=0.7cm, label= below: {}, label=above: {}] (G) at (1,-3) {5};
\node[draw, circle,minimum size=0.7cm, label= right: , label=above: {}] (H) at (3,-3) {6};

\draw (A)  -- node [sloped, anchor=center, above, text width=2.5cm] {} (B);
\draw (A)  -- node [sloped, anchor=center, above, text width=2.5cm] {} (C);
\draw (B)  -- node [sloped, anchor=center, above, text width=2.5cm] {} (E);
\draw (B)  -- node [sloped, anchor=center, above, text width=2.5cm] {} (F);
\draw (C)  -- node [sloped, anchor=center, above, text width=2.5cm] {} (G);
\draw (C)  -- node [sloped, anchor=center, above, text width=2.5cm] {} (H);
\end{tikzpicture}}
\caption{Depth-First-Search (DFS)} \label{DFSsearch}
\end{minipage}
\end{figure}

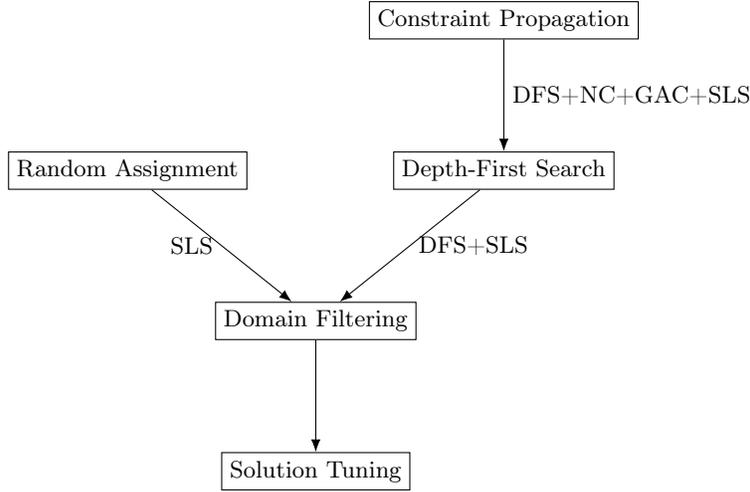
\begin{figure}[ht]
\begin{center}
\begin{tikzpicture}
\node[draw] (A) at (0,7) {Depth-First Search};
\node[draw] (B) at (-5,7) {Random Assignment};
\node[draw] (C) at (-2.5,5) {Domain Filtering};
\node[draw] (D) at (-2.5,3) {Solution Tuning};
\node[draw] (E) at (0,9) {Constraint Propagation};

\draw (B)  -- node[left] {SLS} (C);
\draw (C)  -- (D);
\draw (A)  -- node[right] {DFS+SLS} (C);
\draw (E)  -- node[right] {DFS+NC+GAC+SLS}(A);

\end{tikzpicture}
\caption{SLS-based Solving Approach}
\label{SLSVariants}
\end{center}
\end{figure}

\subsection{Experimentation}

\begin{figure}[ht]
\includegraphics[width=\linewidth]{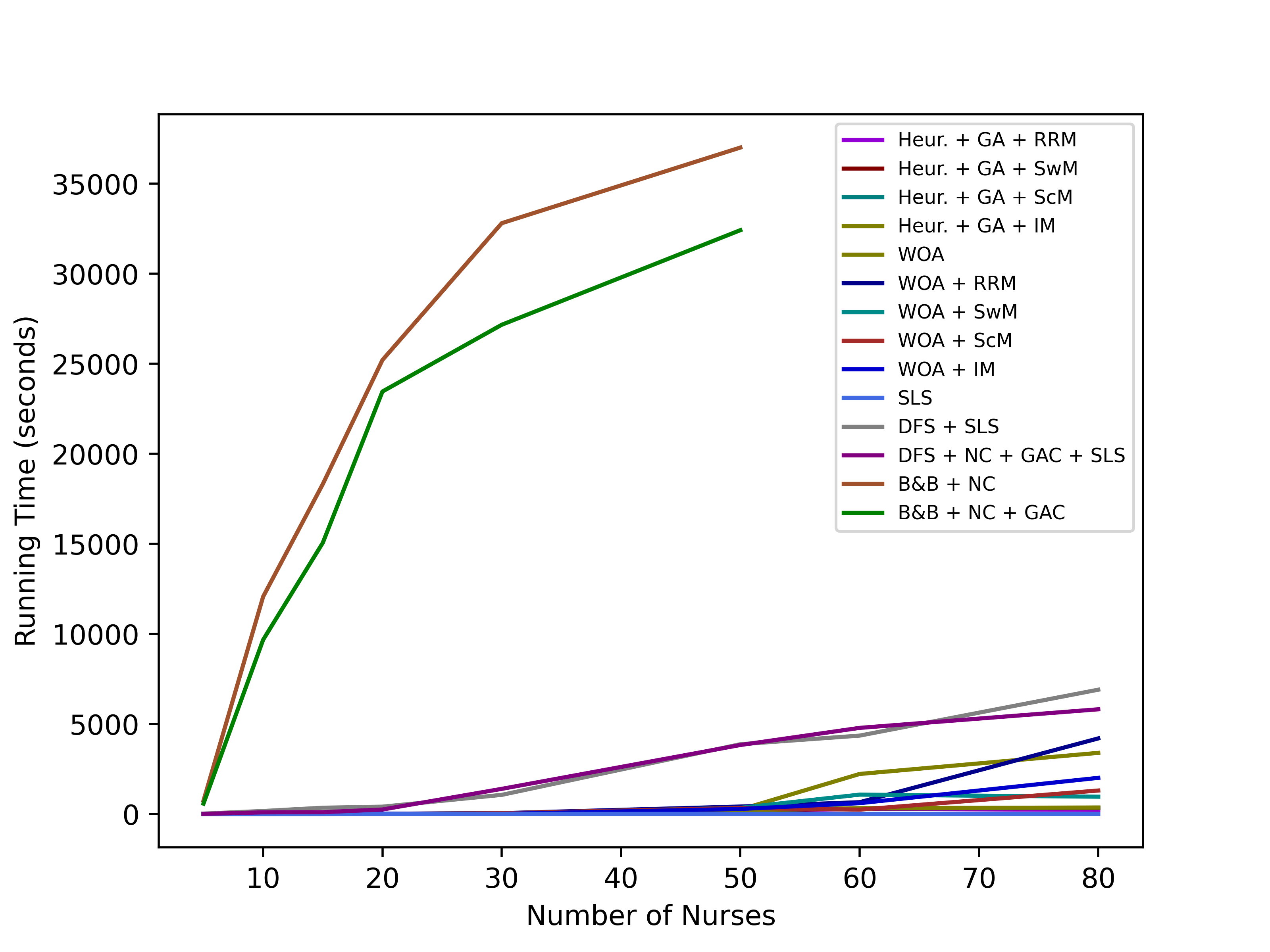}
\caption{Running time of all the methods} \label{NSPRNCurve}
\end{figure}

To asses the efficiency of our proposed methods in practice, and to asses the effect of CP when combined with B\&B and SLS, we have conducted multiple experiments against metaheuristic methods (variants of WOA and hybrid GA methods \cite{icores24}) using multiple NSP instances. The parameters of the NSP instances used in the experiments are presented in Table~\ref{table:12}. The experiments were performed on a personal computer with the following specifications: Intel® Core™ i5-6200U CPU @ 2.3 GHz and 8 GB of RAM. Table~\ref{table:11} presents the experimental results, including the best solutions returned (BS) and their corresponding running times (RT). Additionally, Figure \ref{NSPRNCurve} illustrates the running times for all methods. Our experimentation also involved a solution quality analysis of the approximate methods based on 20 runs, and the result of this analysis covering the best, average, and standard deviation metrics is provided in Table~\ref{table:13}. 

\begin{table}[ht]
\setlength{\tabcolsep}{3.5pt}
\caption{The NSP instances parameters}
\label{table:12}
\centering
\begin{tabular}{|c|c|c|c|c|c|c|} 
 \hline
 \textbf{$n$} & \textbf{$q_{sk}$} & \textbf{$p_{sk}$} & \textbf{$h_i$} & \textbf{$y$} & \textbf{$b_i$}\\
 \hline
 \textbf{5} & 1 & 4 & 5 & 2 & 3 \\ \hline
\textbf{10} & 1 & 7 & 5 & 2 & 3 \\ \hline
\textbf{15} & 1 & 12 & 5 & 2 & 3 \\ \hline
\textbf{20} & 1 & 15 & 5 & 2 & 3 \\ \hline
\textbf{30} & 1 & 25 & 5 & 2 & 3 \\ \hline
\textbf{50} & 1 & 35 & 5 & 2 & 3 \\ \hline
\textbf{60} & 1 & 45 & 5 & 2 & 3 \\ \hline
\textbf{80} & 1 & 65 & 5 & 2 & 3 \\ \hline
\end{tabular}
\end{table}

\begin{table}[ht]
\begin{center}
\caption{Experimental results in various methods for different number of nurses}
\renewcommand\arraystretch{2}
\label{table:11}
\centering
\resizebox{\linewidth}{!}{%
\begin{tabular}{|c|c|c|c|c|c|c|c|c|c|c|c|c|c|c|c|c|} 
 \hline
 \multirow{3}{*}{\textbf{Method}} & \multicolumn{16}{c|}{\textbf{Number of Nurses}} \\
 \cline{2-17}
  & \multicolumn{2}{c|}{\textbf{5}} & \multicolumn{2}{c|}{\textbf{10}} & \multicolumn{2}{c|}{\textbf{15}} & \multicolumn{2}{c|}{\textbf{20}}  & \multicolumn{2}{c|}{\textbf{30}} & \multicolumn{2}{c|}{\textbf{50}} & \multicolumn{2}{c|}{\textbf{60}} & \multicolumn{2}{c|}{\textbf{80}}  \\
 \cline{2-17}
  & \textbf{BS} & \textbf{RT} & \textbf{BS} & \textbf{RT} & \textbf{BC} & \textbf{RT} & \textbf{BS} & \textbf{RT} & \textbf{BS} & \textbf{RT} & \textbf{BS} & \textbf{RT} & \textbf{BS} & \textbf{RT} & \textbf{BS} & \textbf{RT}\\
 \hline
\makecell{\textbf{GA + RRM}} & 8.87 & 3.85 & 16.95 & 6.77 & 22.92 & 8.32 & 34.74 & 11.15 & 49.95 & 21.95 & 88.84 & 150.66 & \textbf{114.38} & 285.57 & \textbf{156.06} & 225.70  \\
\hline
\makecell{\textbf{GA + SwM}} & 9.01 & 6.31 & 17.06 & 6.82 & 24.58 & 8.45 & 35.19 & 11.57 & 51.2 & 22.01 & 89.12 & 189.72 & 114.99 & 298.75 & 156.89 & 331.64 \\
\hline
\makecell{\textbf{GA + ScM}} & 9.45 & 6.68 & 17.20 & 7.11 & 25.23 & 9.22 & 35.20 & 11.89 & 54.70 & 22.48 & 89.65 & 214.43 & 115.58 & 312.71 & 157.69 & 350.14 \\
\hline
\makecell{\textbf{GA + IM}} & 9.51 & 7.03 & 17.25 & 8.32 & 25.64 & 10.18 & 35.43 & 12.89 & 54.97 & 22.84 & 90.15 & 205.98 & 116.11 & 322.44 & 158.67 & 358.15 \\
\hline
\textbf{WOA} & 10.22 & 1.01 & 21.29 & 1.62 & 33.39 & 3.03 & 40.83 & 2.90 & 69.93 & 37.86 & 106.25 & 244.34 & 124.75 & 2221.78 & 177.66 & 3393.23 \\
\hline
\makecell{\textbf{WOA + RRM}} & 10.82 & 1.50 & 21.72 & 1.46 & 30.56 & 1.50 & 43.09 & 9.98 & 65.04 & 44.33 & 105.04 & 412.05 & 129.19 & 653.03 & 170.18 & 4196.58  \\
\hline
\makecell{\textbf{WOA + SwM}} & 10.29 & 0.95 & 22.81 & 1.53 & 32.48 & \textbf{0.97} & 41.85 & 14.96 & 65.28 & 24.40 & 103.26 & 340.61 & 127.09 & 1073.41 & 175.55 & 959.06 \\
\hline
\makecell{\textbf{WOA + ScM}} & 9.78 & 1.73 & 19.56 & \textbf{0.11} & 29.51 & 3.49 & 40.68 & 6.01 & 63.38 & 42.05 & 102.01 & 332.04 & 123.40 & 240.24 & 169.48 & 1301.61 \\
\hline
\makecell{\textbf{WOA + IM}} & 10.45 & 1.85 & 22.19 & 0.21 & 33.01 & 1.60 & 41.21 & 15.41 & 63.49 & 4.14 & 103.78 & 277.68 & 127.66 & 599.18 & 173.49 & 2005.55 \\
\hline
\textbf{SLS} & 11.57 & \textbf{0.69} & 24.10 & 0.50 & 32.99 & 1.23 & 47.28 & \textbf{1.12} & 72.32 & \textbf{1.17} & 109.49 & \textbf{1.49} & 142.92 & \textbf{1.97} & 189.89 & \textbf{2.79} \\
\hline
\makecell{\textbf{DFS + SLS}} & 14.86 & 18.90 & 28.68 & 164.11 & 38.91 & 345.57 & 43.62 & 404.75 & 76.08 & 1060.31 & 119.21 & 3873.79 & 148.34 & 4351.57 & 189.96 & 6901.90 \\
\hline
\makecell{\textbf{DFS + NC + GAC + SLS}} & 12.34 & 5.49 & 25.81 & 89.24 & 32.28 & 100.16 & 49.33 & 246.71 & 67.98 & 1394.28 & 109.96 & 3830.05 & 140.86 & 4781.31 & 185.54 & 5813.59 \\
\hline
\makecell{\textbf{B\&B + NC}} & \textbf{7.89} & 734.69 & \textbf{16.04} & 12071.14 & \textbf{21.53} & 18320.91 & \textbf{30.95} & 25214.34 & \textbf{48.33} & 32814.27 & \textbf{86.98} & 37014.83 & - & - & - & - \\
\hline
\makecell{\textbf{B\&B + NC + GAC}} & \textbf{7.89} & 574.13 & \textbf{16.04} & 9675.23 & \textbf{21.53} & 15060.17 & \textbf{30.95} & 23463.81 & \textbf{48.33} & 26172.43 & \textbf{86.98} & 32426.29 & - & - & - & - \\
\hline
\end{tabular}}
\end{center}
\end{table}


\begin{table}[ht]
\begin{center}
\caption{The Experimental results for the Best, Average, and Standard deviations}
\renewcommand\arraystretch{2}
\label{table:13}
\centering
\resizebox{\linewidth}{!}{%
\begin{tabular}{|c|c|c|c|c|c|c|c|c|c|c|c|c|c|c|c|c|c|c|c|c|c|c|c|c|} 
 \hline
 \multirow{3}{*}{\textbf{Method}} & \multicolumn{24}{c|}{\textbf{Number of Nurses}} \\
 \cline{2-25}
  & \multicolumn{3}{c|}{\textbf{5}} & \multicolumn{3}{c|}{\textbf{10}} & \multicolumn{3}{c|}{\textbf{15}} & \multicolumn{3}{c|}{\textbf{20}}  & \multicolumn{3}{c|}{\textbf{30}} & \multicolumn{3}{c|}{\textbf{50}} & \multicolumn{3}{c|}{\textbf{60}} & \multicolumn{3}{c|}{\textbf{80}}  \\
 \cline{2-25}
  & \textbf{Best} & \textbf{Ave.} & \textbf{Dev.} & \textbf{Best} & \textbf{Ave.} & \textbf{Dev.} & \textbf{Best} & \textbf{Ave.} & \textbf{Dev.} & \textbf{Best} & \textbf{Ave.} & \textbf{Dev.} & \textbf{Best} & \textbf{Ave.} & \textbf{Dev.} & \textbf{Best} & \textbf{Ave.} & \textbf{Dev.} & \textbf{Best} & \textbf{Ave.} & \textbf{Dev.} & \textbf{Best} & \textbf{Ave.} & \textbf{Dev.}\\
 \hline
 \makecell{\textbf{GA + RRM}} & \textbf{8.87} & 10.12 & 0.95 & \textbf{16.95} & 18.12 & 1.12 & \textbf{22.92} & 24.01 & 1.42 & \textbf{34.74} & 36.21 & 1.39 & \textbf{49.95} & 51.54 & 1.51 & \textbf{88.84} & 90.27 & 1.61 & \textbf{114.38} & 116.30 & 1.67 & \textbf{156.06} & 158.39 & 1.49\\
 \hline
 \makecell{\textbf{GA + SwM}} & 9.01 & 10.28 & 1.09 & 17.06 & 18.53 & 1.53 & 24.58 & 25.93 & 1.42 & 35.19 & 36.33 & 1.54 & 51.2 & 52.56 & 0.61 & 89.12 & 91.29 & 1.13 & 114.99 & 116.81 & 1.08 & 156.89 & 158.84 & 1.01 \\
 \hline
 \makecell{\textbf{GA + ScM}} & 9.45 & 10.37 & 0.73 & 17.20 & 18.79 & 0.64 & 25.23 & 26.39 & 0.59 & 35.20 & 36.77 & 0.46 & 54.70 & 56.22 & 1.23 & 89.65 & 93.02 & 1.75 & 115.58 & 117.11 & 1.32 & 157.69 & 159.55 & 1.18 \\
 \hline
 \makecell{\textbf{GA + IM}} & 9.51 & 10.85 & 0.69 & 17.25 & 18.74 & 0.64 & 25.64 & 27.2 & 1.08 & 35.43 & 37.14 & 1.82 & 54.97 & 56.89 & 1.49 & 90.15 & 93.41 & 1.81 & 116.11 & 118.21 & 1.82 & 158.67 & 161.01 & 1.48 \\
 \hline
 \textbf{WOA} & 10.22 & 11.79 & 0.76 & 21.29 & 22.33 & 0.88 & 33.39 & 35.21 & 1.45 & 40.83 & 42.25 & 1.77 & 69.93 & 71.09 & 1.31 & 106.25 & 109.2 & 1.75 & 124.75 & 126.5 & 1.4 & 177.66 & 179.29 & 0.91 \\
 \hline
 \makecell{\textbf{WOA + RRM}} & 10.82 & 11.28 & 0.69 & 21.72 & 23.01 & 1.93 & 30.56 & 32.15 & 1.2 & 43.09 & 44.98 & 1.38 & 65.04 & 67.25 & 1.39 & 105.04 & 107.34 & 2.18 & 129.19 & 131.35 & 1.67 & 170.18 & 172.71 & 1.48 \\
 \hline
 \makecell{\textbf{WOA + SwM}} & 10.29 & 11.89 & 1.54 & 22.81 & 24.51 & 1.45 & 32.48 & 34.33 & 1.69 & 41.85 & 43.39 & 1.67 & 65.28 & 68.01 & 2.18 & 103.26 & 105.21 & 1.88 & 127.09 & 129.92 & 2.22 & 175.55 & 178.18 & 1.49 \\
 \hline
 \makecell{\textbf{WOA + ScM}} & 9.78 & 11.15 & 1.4 & 19.56 & 21.57 & 1.27 & 29.51 & 31.35 & 2.13 & 40.68 & 42.59 & 1.69 & 63.38 & 65.83 & 1.86 & 102.01 & 105.2 & 1.12 & 123.40 & 125.35 & 2.58 & 169.48 & 171.96 & 1.46 \\
 \hline
 \makecell{\textbf{WOA + IM}} & 10.45 & 11.87 & 1.58 & 22.19 & 23.94 & 1.88 & 33.01 & 35.13 & 1.42 & 41.21 & 43.84 & 1.38 & 63.49 & 65.42 & 1.04 & 103.78 & 105.86 & 2.27 & 127.66 & 129.63 & 1.84 & 173.49 & 175.83 & 1.14 \\
 \hline
\textbf{SLS} & 11.57 & 14.44 & 1.33 & 24.10 & 26.81 & 1.48 & 32.99 & 38.09 & 2.28 & 47.28 & 51.45 & 2.03 & 72.32 & 77.75 & 2.49 & 109.49 & 122.74 & 4.67 & 142.92 & 147.69 & 3.67 & 189.89 & 202.84 & 6.91\\
 \hline
\end{tabular}}
\end{center}
\end{table}

The experimental results are very promising as they reveal the efficiency of B\&B in returning the optimal solution, while SLS offers a reasonable trade-off between the solution quality and the running time. Although the experimentation shows remarkable running time improvements when deploying CP (NC and GAC) as a preprocessing step for B\&B and the DFS-based SLS variant, it is fair to note that B\&B still suffer from its exponential time costs (as illustrated in Figure \ref{NSPRNCurve}), and fail in providing a solution for the instances with 60 and 80 nurses. While CP helps reduce the running time for B\&B and DFS, these methods still cannot compete with the approximate methods in this aspect. In conclusion, B\&B outperform all the approximate methods in terms of quality of solution but comes with an expensive time cost, while the approximate methods including SLS trade the quality of the solutions over better running times for all NSP instances. To compensate for the manual modeling and the uncertainty associated with our explicit and implicit approaches, respectively, we propose exact and approximate methods to passively learn the NSP constraints.

\section{Data-driven Learning for the NSP}
\label{learning}
Constraint Programming (CP) and Operation Research (OR) focus on solving real-world combinatorial optimization problems, where the modeling task is fundamental to both fields. Modeling a given combinatorial optimization problem involves representing the constraints that must be satisfied, as well as the objectives to be optimized. These constraints and objectives are then provided to a solver to find a solution a given problem instance. A major challenge in solving optimization problems, lies in the modeling phase, because the method used in the modeling phase can influence the solving process (such as selecting the solver). In general, constraints are often actively and manually learned from domain experts or software systems. This learning process may also involve further verification against relevant contract agreements and past solutions, which may be tedious due to the huge volume of data available that may need to be analyzed. Additionally, obtaining constraints from hospitals may be challenging in practice due to security and privacy limitations. All these facts motivates the automatic modeling approach where the learning of the constraints may be done actively or passively using historical data (past solutions). The challenge with the ML implicit approach we have proposed in Section \ref{implicit} is that it comes with uncertainty related to the quality of the solution, as the implicitly learned patterns do not concretely reveal the satisfied constraints and the optimized objectives. Furthermore, there is no practical method to quantify this uncertainty without explicit knowledge of the constraints. In contrast, our CSP approach (see Section \ref{explicitapproach}) assumes that the constraints and objectives are explicitly defined within the model. So in order to use our explicit solving approach and be able to employ our CSP solvers (B\&B and SLS), we propose a passive learning approach that relies on historical data to learn CSP model, without requiring any user interaction. Furthermore, we investigate the Non-Negative Matrix Factorization (NMF) method to implicitly learn the constraints and objectives (such as nurses preferences). The NMF method works by factorizing a past scheduling solution into a product of two matrices (representing the NSP constraints and preferences) that may be used to predict missing entries in partial scheduling solutions.

\subsection{Passive Learning of CSP model via Matrix Slicing}
\label{WCSPlearningmodel}

Our approach to learning a CSP model is based on the assumption that the constraints to be learned are identified, but their bounds, as well as whether they are satisfied, remain unknown. A typical real-world scenario would be collecting the set of constraints applied in a hospital during a specific planning period. In this context, we adopt the NSP constraints related to the WCSP formulation from \cite{NSPCSP}. The objective of our passive matrix slicing learning method is to determine the bounds of the constraints and also learn whether certain constraints hold within historical data. Consequently, the learned constraints that constitute a CSP model can be fed as an input to our explicit solving approach, and solved using our B\&B and SLS solvers. To achieve this goal, we assume that the historical data is represented as a 2D array (i.e. matrix), where the rows correspond to nurses, the columns represent the planning horizon, and the array entries are binary entries where 1 indicates a nurse is assigned to a shift, and 0 indicates they are not as illustrated in Table \ref{learnchedule}. The CSP model can be learned as follows; the number of nurses corresponds to the number of rows, and the domain is the set of all possible shift patterns based on the number of columns. The domain size is is $2^{sk}$, where k is the number of days and s is the number of shifts per day. To learn the CSP constraints, various operations are performed on different matrix slices to extract the constraints bounds.

\begin{table}[ht]
\centering
\caption{Example of a scheduling solution} 
\label{learnchedule}
\resizebox{7cm}{!}{%
\begin{tabular}{ | c | c | c| c| c| c | c | c | c | c | c |}
    \hline
    & $Day_1Shift_1$ & $Day_1Shift_2$ & $Day_1Shift_3$ & $Day_1Shift_4$ & $...$ & $Day_7Shift_4$\\[1ex] \hline
    $Nurse_1$ & 0 & 1 & 0 & 1 & ... & 1\\[1ex] \hline
    $Nurse_2$ & 1 & 0 & 1 & 0 & ... & 0\\[1ex] \hline
    $Nurse_3$ & 1 & 1 & 0 & 1 & ... & 0\\[1ex] \hline
    $Nurse_4$ & 0 & 1 & 1 & 1 & ... & 1\\[1ex] \hline
    $Nurse_5$ & 0 & 0 & 1 & 1 & ... & 1\\[1ex] \hline
\end{tabular}}
\end{table}

While our explicit learning method is capable of learning a CSP with a large set of constraints from different hospitals, for simplicity, we adopt the NSP constraints related to the WCSP formulation from \cite{NSPCSP} and assume to have four shifts per day. Constraint c2, ``Each schedule must satisfy the hospital’s minimum daily shift coverage requirements" indicates the minimum shift requirement, learning the bound of this constraint is achieved by aggregating all the rows into a 1D array and performing a linear search for the minimum value, which represents this bound. This process may also be used to determine the maximum shift requirement bound by finding for the maximum value in the aggregated sum list. Constraint c3, ``Each nurse may work a maximum number of shifts in a single day" represent the upper bound of number of shifts a nurse can work in each day. This upper bound is learned by finding the number of shifts each nurse worked daily across the schedule, and then identifying the maximum value. The same procedure may be applied to learn the lower bound of number of shifts a nurse can work in each day (excluding off days). Constraint c4, ``No nurse may work a night shift followed immediately by a morning shift'', indicates whether a nurse can work a morning shift immediately following a night shift. Learning this constraint involves verifying whether it was considered in constructing a schedule rather than learning specific bounds, we use a boolean variable to check whether the constraint was satisfied or violated in historical data. Constraint c5, ``Each nurse may not exceed the maximum number of shifts per week", sets an upper bound for the weekly number of working shifts per nurse, and it is learned by summing up the total number of shifts worked by each individual nurse throughout the schedules, and then identifying the highest value among them. This procedure may also be applied to learn the lower bound by identifying the lowest value. The CSP learning process is detailed in Algorithm \ref{algCSP}. Given historical data (a collection of past schedules), the initial step is to identify the dimensions of a scheduling solution which determines the number of nurses and the domain size (assuming all schedules have consistent dimensions). The second step is to learn the set of constraints (including the bounds and the satisfiability of the constraints) by iterating through all the schedules (learning the constraints from a year's worth of schedules would involve 52 schedules). Note that learning a WCSP for a combinatorial optimization problem may require additional information such as the shift patterns' costs/weights related to the domain values for the nurses. Thus, calculating these weights may be done by considering externally provided shift costs for each nurse. For instance, the overall weight for the domain value ``0100100011000000100000101000'' would be 9 if the individual shift costs are as follows; ``shift1: 1'', ``shift2: 2'', ``shift3: 1'', and ``shift4: 3''. Note that in a combinatorial optimization problem where the objective is to optimize nurses' preferences, the costs are assumed to represent the nurses' preferences to work in the individual shifts.

\begin{algorithm}[ht]
\SetAlgoLined
$CSP = \{\emptyset$\}\\
$schedules = \{$1,...,w$\}$\\
$m, n = schedules[1].shape()$\\
$Variables = \{$1,...,m$\}$\\
$Domain = \{$1,...,$2^{n}$$\}$\\
$c2 =\infty$\\
$c3 = c5 =0$\\
$c4 =True$\\
$ConstraintsLearner(schedules):$\\
\hspace{0.1 in}${\bf for} \ i \ {\bf in} \ range(0, w):$\\
\hspace{0.2 in}$tmp1 \leftarrow MinShiftRequirements(schedules[i])$\\
\hspace{0.2 in}${\bf if} \ tmp1 < c2:$\\
\hspace{0.3 in}$c2 = tmp1$\\
\hspace{0.2 in}$tmp2 \leftarrow MaxShiftsperDay(schedules[i])$\\
\hspace{0.2 in}${\bf if} \ tmp2 > c3:$\\
\hspace{0.3 in}$c3 = tmp2$\\
\hspace{0.2 in}$tmp3 \leftarrow NightMorningShifts(schedules[i])$\\
\hspace{0.2 in}${\bf if} \ tmp3 = False:$\\
\hspace{0.3 in}$c4 = False$\\
\hspace{0.2 in}$tmp4 \leftarrow MaxShiftsperWeek(schedules[i])$\\
\hspace{0.2 in}${\bf if} \ tmp4 > c5:$\\
\hspace{0.3 in}$c5 = tmp4$\\
$Constraints = \{$c2, c3, c4, c5$\}$\\
$CSP \leftarrow \{Variables, Domain, Constraints\}$\\
{\bf Return}  $CSP$
\caption{The CSP learning process}
\label{algCSP}
\end{algorithm}

One of the challenges in learning a CSP model in the context of NSP, is the large domain size that is formed by all possible value combinations over the planning horizon. For example, if the planning horizon covers 7 days with 4 shifts per day, the domain would consist of $2^{28}$ shift patterns. For this reason, CP techniques \cite{5701859,dechter2003constraint,lecoutre2006generalized,cheng2010mdd,Larrosa} are considered to enforce local consistency before solving to minimize the domain size (see Section \ref{CP}).

\subsection{Experimentation}
To evaluate the performance of our explicit learning method, we conduct an experiment and asses how it performs in terms of running time, when learning constraints from from varying numbers of schedules. In this context, we rely on historical data from the NSPLib library \cite{lib} that provides several weekly NSP coverage requirements for 25 nurses. The results of our experiment are reported in Table \ref{csplearningtable}, and they demonstrate that our method is capable of learning NSP constraints from different numbers of schedules, and the running time proportionally increases as the number of schedules grow. Since it may be challenging for our exact learning method to perform well in terms of running time when using a large number of schedules (especially in the case of learning a large number of constraints), we propose an alternative approximate method relying on ML that works by factorizing past solutions into a set of two matrices, representing the constraints as well as nurses' preferences.

\begin{table}[ht]
\centering
\caption{Running time of learning the CSP model}
\label{csplearningtable}
\begin{tabular}{|c|c|}
\hline
Nº of Schedules & Running Time (s) \\
\hline
10 & 1.07 \\ \hline
20 & 2.19 \\ \hline
30 & 3.37 \\ \hline
40 & 4.62 \\ \hline
50 & 5.91 \\ \hline
60 & 7.95 \\ \hline
70 & 8.65 \\ \hline
80 & 9.22 \\ \hline
90 & 9.88 \\ \hline
100 & 11.54 \\ \hline
\end{tabular}
\end{table}

\subsection{{Non-Negative Matrix Factorization (NMF)}}
\label{NMFsection}
Non-Negative Matrix Factorization (NMF) is an unsupervised ML technique mainly applied in topic modeling and dimensionality reduction tasks \cite{secrypt22,Kuang2014NonnegativeMF}. NMF may also be leveraged in various fields including recommender systems, signal processing, document clustering, etc. The goal of NMF is to decompose multivariate data into a user-specified number of features. Given a non-negative matrix X, NMF decomposes X into two lower-rank non-negative matrices H and W. The product of H and W serves as an approximation of X ($X \approx W \cdot H$). Consider a rank-k matrix X of dimensions $m \times n$ with no negative entries. There exist two rank-k matrices W and H of dimensions $m \times r$ and $r \times n$, respectively, where r is smaller than both m and n). Note that k is the number of features that may be set by the user. For instance, in the context of NSP, k correspond to the number of shifts per day. Finding an accurate approximation of X using NMF involves solving a minimization problem. NMF works by initially setting matrices H and W with random values, and then iteratively tuning these matrices to minimize the distance between their product and X relying on a distance metric. This process uses the multiplicative update equations \cite{10.1162/NECO_a_00168} as illustrated in Equation \ref{Hequation} and Equation \ref{Wequation}, respectively. Finally, the algorithm stops once the average error (with respect to the distance metric) converges, or once a maximum number of iterations is reached. 

\begin{equation}
\label{Hequation}
H_{ij}^{n+1} \leftarrow H_{ij}^{n} \cdot 
\frac
{((W^n)^T X)_{ij}}
{((W^n)^T W^n H^n)_{ij}}
\end{equation}

\begin{equation}
\label{Wequation}
W_{ij}^{n+1} \leftarrow W_{ij}^{n} \cdot
\frac
{(X (H^{n+1})^T)_{ij}}
{(W^n H^{n+1} (H^{n+1})^T)_{ij}}
\end{equation}

To evaluate the quality of NMF factorization, an error function needs to be used to measure how accurately the approximation represents the original data. Therefore, the Frobenius Norm is used to quantify the error rate at each iteration of the algorithm, as shown in Equation \ref{FNeq2}. The stopping criterion is determined either when a specified number of iterations is reached or when the quantified FN error convergences.

\begin{equation}
\label{FNeq2}
Minimize(\lVert{X_{ij} - (W H)_{ij}}\lVert_F)
\end{equation}

\begin{exmp}

To illustrate how NMF is applies to learn the NSP constraints and preferences, we use a scheduling example that shows the number of worked days during a week, as shown in Table \ref{NMFinput}. NMF requires the entries in the input matrix to be non-negative, which is satisfied in the context of NSP, since each data entry correspond to a positive value indicating the count of worked days per week.

\begin{table}[ht]
\centering
\caption{A week scheduling example (X)} 
\label{NMFinput}
\resizebox{7cm}{!}{%
	\begin{tabular}{ | c | c | c| c| c| c | c | c | c | c |}
 		\hline
 	& $Day_1$ & $Day_2$ & $Day_3$ & $Day_4$ & $Day_5$&$Day_6$ & $Day_7$ \\[1ex] \hline
 	$Nurse_1$ & 2 & 1 & 1 & 3 & 1 & 0 & 2\\[1ex] \hline
		$Nurse_2$ & 4 & 0 & 3 & 1 & 1 & 2 & 3\\[1ex] \hline
		$Nurse_3$ & 1 & 2 & 2 & 3 & 1 & 0 & 1\\[1ex] \hline
		$Nurse_4$ & 2 & 2 & 1 & 1 & 0 & 3 & 1\\[1ex] \hline
		$Nurse_5$ & 3 & 1 & 2 & 0 & 4 & 1 & 1\\[1ex] \hline
	\end{tabular}
 }
 \end{table} 

The resulting W and H matrices after applying NMF are shown in Table \ref{HandW}, where $Shift_1$, $Shift_2$, and $Shift_3$ are three features that captures the quantities of interest concerning nurses' preferences for working those shifts. As indicated above, matrices W and H are expected to approximate the original matrix X as accurate as possible such that X can be reconstructed by calculating the matrix product of W and H. The approximated matrices W and H cannot be used to fill out missing entries for every matrix, because some specifications must be satisfied to utilize the approximated matrices for predictions. Therefore, in order to use them, the partially filled matrix must have the same dimensions as matrix X, and also suggest a threshold related to whether or not a given matrix can be used for prediction. This threshold can be the FN error distance between matrix X and the product of W and H. So if the FN error between the new partially filled matrix and matrix X satisfy this threshold, the new matrix is may be used for prediction, otherwise, it cannot.

\begin{table}[ht]
\centering
\caption{Constraints (H) and Preferences (W)}
\label{HandW}
\begin{minipage}{0.45\textwidth}
\centering
\resizebox{7cm}{!}{%
\begin{tabular}{ | c | c | c| c| c| c | c | c | c | c | c | }
\hline
&$Day_1$ & $Day_2$ & $Day_3$ & $Day_4$ & $Day_5$&$Day_6$ & $Day_7$ \\[1ex] \hline
$Shift_1$ & 1.71 & 0.21 & 0.99 & 0.07  & 0 & 1.66  & 1.14 \\[1ex] \hline
$Shift_2$ & 1.19 & 0.27 & 0.88  & 0 & 1.97 & 0 & 0.37 \\[1ex] \hline
$Shift_3$ & 0.53 & 1.16 & 0.76  & 2.41 & 0 & 0 & 0.89 \\[1ex] \hline
\end{tabular}}
\end{minipage}%
\hfill
\begin{minipage}{0.45\textwidth}
\centering
\begin{tabular}{| c | c |c |c |}
 		\hline
      & $Shift_1$ & $Shift_2$ & $Shift_3$ \\[1ex] \hline
    $Nurse_1$ & 0.22 & 0.51 & 1.17 \\[1ex] \hline
    $Nurse_2$ & 1.65 &  0.67 & 0.35 \\[1ex] \hline
    $Nurse_3$ & 0 & 0.53  & 1.30 \\[1ex] \hline
    $Nurse_4$ & 1.21  &  0 & 0.42 \\[1ex] \hline
    $Nurse_5$ & 0.43  &  1.98 & 0 \\[1ex] \hline
\end{tabular}
\end{minipage}
\end{table}
\end{exmp}

\section{Conclusion and Future Work}
To conclude, given the challenge that comes from modeling the NSP manually or automatically, we have proposed an implicit solving approach relying on ML methods and historical data without any prior knowledge. In addition to evaluating the ML methods individually, we further conducted an experiment against an explicit method named COUNT-OR which learns constraints from historical data based on predefined metrics and then applies them to generate new solutions. The experiment involved quantifying the average error by calculating the distance between the generated solutions and historical data using the Frobenius Norm, which demonstrated that our methods outperformed COUNT-OR. Given that our proposed ML implicit approach comes with uncertainly related to the quality of the solution, we have proposed an explicit solving approach to model and solve the NSP relying on the CSP framework. In this context, we have formulated the NSP as a WCSP and further proposed new variants of B\&B and SLS methods as exact and approximate solvers, respectively. Furthermore, we have applied CP techniques to minimize the exponential time cost related our solvers (B\&B in particular). To evaluate the performance of our proposed solving methods, we considered additional metaheuristic methods, and conducted an experiment to evaluate the quality of the solutions and the running time for each method. While the experimental results demonstrated that the approximate methods, particularly SLS, provide a good trade-off between solution quality and the corresponding running time, the results also show that B\&B still suffer from the exponential running time regardless of applying CP. In addition to solving the NSP implicitly and explicitly, we have also tackled the limitations that comes from manual modeling, by proposing two learning methods that passively and automatically learn NSP constraints from historical data. These two learning methods may serve as an initial step to model the NSP for our explicit solving approach. Given the promising results obtained in scope of this paper, in the near future, we plan to consult with  hospitals to evaluate our proposed solving approaches using real-world NSPs. Moreover, we plan to develop nature-inspired techniques to solve NSPs, modeled as a WCSP \cite{talbi2009metaheuristics,RePEc:spr:snopef:v:2:y:2021:i:3:d:10.1007_s43069-021-00068-x,DBLP:conf/smc/BidarM19}. Additionally, we plan to investigate an NSP variant that involve ordinal nurses' preferences by extending our WCSP model to conditional qualitative preferences \cite{DBLP:conf/smc/AlkhiriM21}. Furthermore, since fairness is a significant part of solving the NSP, we plan to address it more precisely by relying on the Multi-objective optimization framework to model additional constraints and objectives that maximize nurse preferences while ensuring equal workload distribution. Finally, we will also investigate tackling NSPs in a dynamic environment, where schedules need to be adjusted in real-time, due to unexpected changes in demand or sudden nurse absences (the COVID-19 pandemic is a good example of such scenarios). Building on previous research work \cite{DBLP:journals/orf/BidarM22,mouhoub2003arc}, the target is to develop a solving approach that can generate new scheduling solutions with minimal perturbation (in terms of re-assignments) within an acceptable timeframe.

%
%
%
%
\bibliographystyle{ieeetr}
\bibliography{mybibliography}

\begin{thebibliography}{10}

\bibitem{woeginger2003exact}
G.~J. Woeginger, ``Exact algorithms for np-hard problems: A survey,'' in {\em Combinatorial optimization—eureka, you shrink!}, pp.~185--207, Springer, 2003.

\bibitem{RePEc:spr:snopef:v:2:y:2021:i:3:d:10.1007_s43069-021-00068-x}
W.~Korani and M.~Mouhoub, ``{Review on Nature-Inspired Algorithms},'' {\em SN Operations Research Forum}, vol.~2, pp.~1--26, September 2021.

\bibitem{B}
A.~Passerini, G.~Tack, and T.~Guns, ``Introduction to the special issue on combining constraint solving with mining and learning,'' {\em Artificial Intelligence}, vol.~244, pp.~1--5, 03 2017.

\bibitem{NSPML}
A.~Ben~Said, E.~A. Mohammed, and M.~Mouhoub, ``An implicit learning approach for solving the nurse scheduling problem,'' in {\em Neural Information Processing} (T.~Mantoro, M.~Lee, M.~A. Ayu, K.~W. Wong, and A.~N. Hidayanto, eds.), (Cham), pp.~145--157, Springer International Publishing, 2021.

\bibitem{NSPCSP}
A.~Ben~Said and M.~Mouhoub, ``A constraint satisfaction problem (csp) approach for the nurse scheduling problem,'' in {\em 2022 IEEE Symposium Series on Computational Intelligence (SSCI)}, pp.~790--795, 2022.

\bibitem{10.1007/978-3-031-34020-8_5}
M.~Sadeghilalimi, M.~Mouhoub, and A.~B. Said, ``Solving the nurse scheduling problem using the whale optimization algorithm,'' in {\em Optimization and Learning} (B.~Dorronsoro, F.~Chicano, G.~Danoy, and E.-G. Talbi, eds.), (Cham), pp.~62--73, Springer Nature Switzerland, 2023.

\bibitem{icores24}
M.~Sadeghilalimi., M.~Mouhoub., and A.~{Ben Said}., ``Evolutionary techniques for the nurse scheduling problem,'' in {\em Proceedings of the 13th International Conference on Operations Research and Enterprise Systems - ICORES}, pp.~333--340, INSTICC, SciTePress, 2024.

\bibitem{maenhout2010branching}
B.~Maenhout and M.~Vanhoucke, ``Branching strategies in a branch-and-price approach for a multiple objective nurse scheduling problem,'' {\em Journal of scheduling}, vol.~13, no.~1, pp.~77--93, 2010.

\bibitem{abdennadher1999nurse}
S.~Abdennadher and H.~Schlenker, ``Nurse scheduling using constraint logic programming,'' in {\em AAAI/IAAI}, pp.~838--843, 1999.

\bibitem{jaumard1998generalized}
B.~Jaumard, F.~Semet, and T.~Vovor, ``A generalized linear programming model for nurse scheduling,'' {\em European journal of operational research}, vol.~107, no.~1, pp.~1--18, 1998.

\bibitem{legrain2020rotation}
A.~Legrain, J.~Omer, and S.~Rosat, ``A rotation-based branch-and-price approach for the nurse scheduling problem,'' {\em Mathematical Programming Computation}, vol.~12, no.~3, pp.~417--450, 2020.

\bibitem{jan2000evolutionary}
A.~Jan, M.~Yamamoto, and A.~Ohuchi, ``Evolutionary algorithms for nurse scheduling problem,'' in {\em Proceedings of the 2000 Congress on Evolutionary Computation. CEC00 (Cat. No. 00TH8512)}, vol.~1, pp.~196--203, IEEE, 2000.

\bibitem{gutjahr2007aco}
W.~J. Gutjahr and M.~S. Rauner, ``An aco algorithm for a dynamic regional nurse-scheduling problem in austria,'' {\em Computers \& Operations Research}, vol.~34, no.~3, pp.~642--666, 2007.

\bibitem{wu2013ant}
J.-j. Wu, Y.~Lin, Z.-h. Zhan, W.-n. Chen, Y.-b. Lin, and J.-y. Chen, ``An ant colony optimization approach for nurse rostering problem,'' in {\em 2013 IEEE International Conference on Systems, Man, and Cybernetics}, pp.~1672--1676, IEEE, 2013.

\bibitem{jafari2015maximizing}
H.~Jafari and N.~Salmasi, ``Maximizing the nurses’ preferences in nurse scheduling problem: mathematical modeling and a meta-heuristic algorithm,'' {\em Journal of industrial engineering international}, vol.~11, no.~3, pp.~439--458, 2015.

\bibitem{rajeswari2017directed}
M.~Rajeswari, J.~Amudhavel, S.~Pothula, and P.~Dhavachelvan, ``Directed bee colony optimization algorithm to solve the nurse rostering problem,'' {\em Computational intelligence and neuroscience}, vol.~2017, 2017.

\bibitem{KumarMohit2021DCSP}
M.~Kumar, S.~Kolb, C.~Gautrais, and L.~De~Raedt, ``Democratizing constraint satisfaction problems through machine learning,'' in {\em AAAI}, pp.~16057--16059, 2021.

\bibitem{kumar}
M.~Kumar, S.~Teso, P.~De~Causmaecker, and L.~De~Raedt, ``Automating personnel rostering by learning constraints using tensors,'' in {\em 2019 IEEE 31st International Conference on Tools with Artificial Intelligence (ICTAI)}, pp.~697--704, 2019.

\bibitem{tacle}
S.~Paramonov, S.~Kolb, T.~Guns, and L.~De~Raedt, ``Tacle: Learning constraints in tabular data,'' in {\em Proceedings of the 2017 ACM on Conference on Information and Knowledge Management}, CIKM '17, (New York, NY, USA), p.~2511–2514, Association for Computing Machinery, 2017.

\bibitem{E}
M.~Kumar, S.~Teso, and L.~De~Raedt, ``Acquiring integer programs from data,'' in {\em Proceedings of the Twenty-Eighth International Joint Conference on Artificial Intelligence, {IJCAI-19}}, pp.~1130--1136, 7 2019.

\bibitem{baskaran2014integer}
G.~Baskaran, A.~Bargiela, and R.~Qu, ``Integer programming: Using branch and bound to solve the nurse scheduling problem,'' in {\em 2014 International Conference on Artificial Intelligence and Manufacturing Engineering (IIE ICAIME2014)}, 2014.

\bibitem{burke2001memetic}
E.~Burke, P.~Cowling, P.~De~Causmaecker, and G.~V. Berghe, ``A memetic approach to the nurse rostering problem,'' {\em Applied intelligence}, vol.~15, no.~3, pp.~199--214, 2001.

\bibitem{zhang2011hybrid}
Z.~Zhang, Z.~Hao, and H.~Huang, ``Hybrid swarm-based optimization algorithm of ga \& vns for nurse scheduling problem,'' in {\em International Conference on Information Computing and Applications}, pp.~375--382, Springer, 2011.

\bibitem{I}
A.~A. Constantino, E.~L. de~Melo, D.~Landa-Silva, and W.~Romão, ``A heuristic algorithm for nurse scheduling with balanced preference satisfaction,'' in {\em 2011 IEEE Symposium on Computational Intelligence in Scheduling (SCIS)}, pp.~39--45, 2011.

\bibitem{J}
A.~Aparecido~Constantino, E.~Tozzo, R.~Lankaites~Pinheiro, D.~Landa-Silva, and W.~Rom\~{a}o, ``A variable neighbourhood search for nurse scheduling with balanced preference satisfaction,'' in {\em Proceedings of the 17th International Conference on Enterprise Information Systems - Volume 1}, p.~462–470, SCITEPRESS - Science and Technology Publications, Lda, 2015.

\bibitem{Tassopoulos}
I.~X. Tassopoulos, I.~P. Solos, and G.~N. Beligiannis, ``Α two-phase adaptive variable neighborhood approach for nurse rostering,'' {\em Computers \& operations research}, vol.~60, pp.~150--169, 2015.

\bibitem{YilmazEbru2010AMPM}
E.~Yilmaz, ``A mathematical programming model for scheduling of nurses’ labor shifts,'' {\em Journal of medical systems}, vol.~36, no.~2, pp.~491--496, 2010.

\bibitem{li}
J.~Li and U.~Aickelin, ``A bayesian optimization algorithm for the nurse scheduling problem,'' in {\em The 2003 Congress on Evolutionary Computation, 2003. CEC '03.}, vol.~3, pp.~2149--2156 Vol.3, 2003.

\bibitem{Aickelin_2007}
U.~Aickelin and J.~Li, ``An estimation of distribution algorithm for nurse scheduling,'' {\em Annals of Operations Research}, vol.~155, p.~289–309, Jul 2007.

\bibitem{H}
S.~Karmakar, S.~Chakraborty, T.~Chatterjee, A.~Baidya, and S.~Acharyya, ``Meta-heuristics for solving nurse scheduling problem: A comparative study,'' {\em 2016 2nd International Conference on Advances in Computing, Communication, \& Automation (ICACCA) (Fall)}, pp.~1--5, 2016.

\bibitem{BESSIERE2017315}
C.~Bessiere, F.~Koriche, N.~Lazaar, and B.~O'Sullivan, ``Constraint acquisition,'' {\em Artificial Intelligence}, vol.~244, pp.~315--342, 2017.
\newblock Combining Constraint Solving with Mining and Learning.

\bibitem{DBLP:conf/ijcai/AlanaziMZ16}
E.~Alanazi, M.~Mouhoub, and S.~Zilles, ``The complexity of learning acyclic cp-nets,'' in {\em Proceedings of the Twenty-Fifth International Joint Conference on Artificial Intelligence, {IJCAI} 2016, New York, NY, USA, 9-15 July 2016} (S.~Kambhampati, ed.), pp.~1361--1367, {IJCAI/AAAI} Press, 2016.

\bibitem{DBLP:journals/ai/AlanaziMZ20}
E.~Alanazi, M.~Mouhoub, and S.~Zilles, ``The complexity of exact learning of acyclic conditional preference networks from swap examples,'' {\em Artif. Intell.}, vol.~278, 2020.

\bibitem{mouhoub2023exact}
M.~Mouhoub, H.~Al~Marri, and E.~Alanazi, ``Exact learning of qualitative constraint networks from membership queries,'' {\em International Journal of Software Engineering and Knowledge Engineering}, pp.~1--27, 2023.

\bibitem{BESSIERE2023103896}
C.~Bessiere, C.~Carbonnel, A.~Dries, E.~Hebrard, G.~Katsirelos, N.~Narodytska, C.-G. Quimper, K.~Stergiou, D.~C. Tsouros, and T.~Walsh, ``Learning constraints through partial queries,'' {\em Artificial Intelligence}, vol.~319, p.~103896, 2023.

\bibitem{modelseeker}
N.~Beldiceanu and H.~Simonis, ``A model seeker: Extracting global constraint models from positive examples,'' in {\em Principles and Practice of Constraint Programming} (M.~Milano, ed.), (Berlin, Heidelberg), pp.~141--157, Springer Berlin Heidelberg, 2012.

\bibitem{ceschia2015second}
S.~Ceschia, N.~T.~T. Dang, P.~D. Causmaecker, S.~Haspeslagh, and A.~Schaerf, ``Second international nurse rostering competition (inrc-ii) --- problem description and rules ---,'' 2015.

\bibitem{lirias3973866}
M.~Kumar, S.~Kolb, and T.~Guns, ``Learning constraint programming models from data using generate-and-aggregate,'' 2022-07-23.

\bibitem{F}
H.~Yao, H.~Hamilton, and C.~Butz, ``A foundational approach to mining itemset utilities from databases,'' in {\em Proceedings of the Fourth {SIAM} International Conference on Data Mining, Lake Buena Vista, Florida, USA, April 22-24, 2004}, vol.~4, 04 2004.

\bibitem{10.5555/645920.672836}
R.~Agrawal and R.~Srikant, ``Fast algorithms for mining association rules in large databases,'' in {\em Proceedings of the 20th International Conference on Very Large Data Bases}, VLDB '94, (San Francisco, CA, USA), p.~487–499, Morgan Kaufmann Publishers Inc., 1994.

\bibitem{twophase}
Y.~Liu, W.-k. Liao, and A.~Choudhary, ``A two-phase algorithm for fast discovery of high utility itemsets,'' in {\em Advances in Knowledge Discovery and Data Mining} (T.~B. Ho, D.~Cheung, and H.~Liu, eds.), (Berlin, Heidelberg), pp.~689--695, Springer Berlin Heidelberg, 2005.

\bibitem{lib}
M.~Vanhoucke and B.~Maenhout, ``Nsplib–a nurse scheduling problem library: a tool to evaluate (meta-) heuristic procedures,'' 01 2007.

\bibitem{M}
L.~Vu and G.~Alaghband, ``An efficient approach for mining association rules from sparse and dense databases,'' in {\em 2014 World Congress on Computer Applications and Information Systems (WCCAIS)}, pp.~1--8, 2014.

\bibitem{laplace}
E.~Setyaningsih and I.~Listiowarni, ``Categorization of exam questions based on bloom taxonomy using naïve bayes and laplace smoothing,'' in {\em 2021 3rd East Indonesia Conference on Computer and Information Technology (EIConCIT)}, pp.~330--333, 04 2021.

\bibitem{K}
N.~Friedman, ``A qualitative markov assumption and its implications for belief change,'' in {\em In Proc. Twelfth Conference on Uncertainty in Artificial Intelligence (UAI '96}, pp.~263--273, Morgan Kaufmann, 1996.

\bibitem{L}
A.~Böttcher and D.~Wenzel, ``The frobenius norm and the commutator,'' {\em Linear Algebra and its Applications}, vol.~429, no.~8, pp.~1864--1885, 2008.

\bibitem{Stefan2020ABAf}
S.~Rocktäschel, {\em A Branch-and-Bound Algorithm for Multiobjective Mixed-integer Convex Optimization}.
\newblock Ilmenau, Germany: Springer Spektrum, Wiesbaden, 1st ed. 2020..~ed., 2020.

\bibitem{BruscoMichaelJ2005BAiC}
M.~J. Brusco, {\em Branch-and-Bound Applications in Combinatorial Data Analysis}.
\newblock Statistics and Computing, Springer, New York, NY, 1st ed. 2005..~ed., 2005.

\bibitem{HaouariM2005Stgm}
M.~Haouari, J.~Chaouachi, and M.~Dror, ``Solving the generalized minimum spanning tree problem by a branch-and-bound algorithm,'' {\em The Journal of the Operational Research Society}, vol.~56, no.~4, pp.~382--389, 2005.

\bibitem{dechter2003constraint}
R.~Dechter and D.~Cohen, {\em Constraint processing}.
\newblock Morgan Kaufmann, 2003.

\bibitem{DBLP:conf/appinf/Mouhoub03}
M.~Mouhoub, ``Dynamic path consistency for interval-based temporal reasoning,'' in {\em The 21st {IASTED} International Multi-Conference on Applied Informatics {(AI} 2003), February 10-13, 2003, Innsbruck, Austria} (M.~H. Hamza, ed.), pp.~393--398, {IASTED/ACTA} Press, 2003.

\bibitem{DBLP:journals/apin/YongM18}
K.~W. Yong and M.~Mouhoub, ``Using conflict and support counts for variable and value ordering in csps,'' {\em Appl. Intell.}, vol.~48, no.~8, pp.~2487--2500, 2018.

\bibitem{mouhoub2011heuristic}
M.~Mouhoub and B.~Jafari, ``Heuristic techniques for variable and value ordering in csps,'' in {\em Proceedings of the 13th annual conference on Genetic and evolutionary computation}, pp.~457--464, 2011.

\bibitem{Larrosa}
J.~Larrosa, ``Node and arc consistency in weighted csp,'' in {\em Eighteenth National Conference on Artificial Intelligence}, (USA), p.~48–53, American Association for Artificial Intelligence, 2002.

\bibitem{BessiereR97}
C.~Bessi{\`{e}}re and J.~R{\'{e}}gin, ``Arc consistency for general constraint networks: Preliminary results,'' in {\em Proceedings of the Fifteenth International Joint Conference on Artificial Intelligence, {IJCAI} 97, Nagoya, Japan, August 23-29, 1997, 2 Volumes}, pp.~398--404, Morgan Kaufmann, 1997.

\bibitem{Regin96}
J.~R{\'{e}}gin, ``Generalized arc consistency for global cardinality constraint,'' in {\em Proceedings of the Thirteenth National Conference on Artificial Intelligence and Eighth Innovative Applications of Artificial Intelligence Conference, {AAAI} 96, {IAAI} 96, Portland, Oregon, USA, August 4-8, 1996, Volume 1} (W.~J. Clancey and D.~S. Weld, eds.), pp.~209--215, {AAAI} Press / The {MIT} Press, 1996.

\bibitem{NSPOLA}
M.~Sadeghilalimi, M.~Mouhoub, and A.~Ben~Said, ``Solving the nurse scheduling problem using the whale optimization algorithm,'' in {\em International Conference in Optimization and Learning (OLA2023)}, p.~To appear, 2023.

\bibitem{5701859}
M.~Dib, R.~Abdallah, and A.~Caminada, ``Arc-consistency in constraint satisfaction problems: A survey,'' in {\em 2010 Second International Conference on Computational Intelligence, Modelling and Simulation}, pp.~291--296, 2010.

\bibitem{lecoutre2006generalized}
C.~Lecoutre and R.~Szymanek, ``Generalized arc consistency for positive table constraints,'' in {\em International conference on principles and practice of constraint programming}, pp.~284--298, Springer, 2006.

\bibitem{cheng2010mdd}
K.~C. Cheng and R.~H. Yap, ``An mdd-based generalized arc consistency algorithm for positive and negative table constraints and some global constraints,'' {\em Constraints}, vol.~15, no.~2, pp.~265--304, 2010.

\bibitem{secrypt22}
R.~Osei., H.~Louafi., M.~Mouhoub., and Z.~Zhu., ``Efficient iot device fingerprinting approach using machine learning,'' in {\em Proceedings of the 19th International Conference on Security and Cryptography - SECRYPT}, pp.~525--533, INSTICC, SciTePress, 2022.

\bibitem{Kuang2014NonnegativeMF}
D.~{da Kuang}, J.~Choo, and H.~Park, {\em Nonnegative matrix factorization for interactive topic modeling and document clustering}, pp.~215--243.
\newblock Springer International Publishing, Jan. 2015.
\newblock Publisher Copyright: {\textcopyright} Springer International Publishing Switzerland 2015.

\bibitem{10.1162/NECO_a_00168}
C.~Févotte and J.~Idier, ``{Algorithms for Nonnegative Matrix Factorization with the β-Divergence},'' {\em Neural Computation}, vol.~23, pp.~2421--2456, 09 2011.

\bibitem{talbi2009metaheuristics}
E.-G. Talbi, {\em Metaheuristics: from design to implementation}.
\newblock John Wiley \& Sons, 2009.

\bibitem{DBLP:conf/smc/BidarM19}
M.~Bidar and M.~Mouhoub, ``Solving weighted constraint satisfaction problems using a new self-adaptive discrete firefly algorithm,'' in {\em 2019 {IEEE} International Conference on Systems, Man and Cybernetics, {SMC} 2019, Bari, Italy, October 6-9, 2019}, pp.~2198--2205, {IEEE}, 2019.

\bibitem{DBLP:conf/smc/AlkhiriM21}
H.~Alkhiri and M.~Mouhoub, ``Weighted constrained cp-nets: an extension of constrained cp-nets with weighted constraints,'' in {\em 2021 {IEEE} International Conference on Systems, Man, and Cybernetics, {SMC} 2021, Melbourne, Australia, October 17-20, 2021}, pp.~1685--1690, {IEEE}, 2021.

\bibitem{DBLP:journals/orf/BidarM22}
M.~Bidar and M.~Mouhoub, ``Nature-inspired techniques for dynamic constraint satisfaction problems,'' {\em Oper. Res. Forum}, vol.~3, no.~2, 2022.

\bibitem{mouhoub2003arc}
M.~Mouhoub, ``Arc consistency for dynamic csps,'' in {\em International Conference on Knowledge-Based and Intelligent Information and Engineering Systems}, pp.~393--400, Springer, 2003.

\end{thebibliography}

\end{document}